%% file: mlhc.tex
\documentclass[pmlr]{jmlr}% new name PMLR (Proceedings of Machine Learning)

\usepackage{bbm}
\usepackage{longtable}% for long tables

 \usepackage{booktabs}
 
\usepackage[load-configurations=version-1]{siunitx} % newer version

\makeatletter
\def\set@curr@file#1{\def\@curr@file{#1}} %temp workaround for 2019 latex release
\makeatother

\theorembodyfont{\upshape}
\theoremheaderfont{\scshape}
\theorempostheader{:}
\theoremsep{\newline}

\jmlrvolume{}
\jmlryear{2023}
\jmlrworkshop{Machine Learning for Healthcare}

\title[Unified Multi-modal Data Embedding and Modality-Aware Attention]{Learning Missing Modal Electronic Health Records with Unified Multi-modal Data Embedding and Modality-Aware Attention}
\author{\Name{Kwanhyung Lee$^{1}$}
       \Email{kwanlee9209@aitrics.com} \\ 
       \Name{Soojeong Lee$^{1,3}$\thanks{Work done while S.Lee was an intern at AITRICS Inc.} \Email{drlisa@aitrics.com}\\}
       \Name{Sangchul Hahn$^{1}$}
       \Email{s.hahn@aitrics.com} \\ 
       \Name{Heejung Hyun$^{1}$}
       \Email{alex.hyun@aitrics.com} \\ 
       \Name{Edward Choi$^{2}$}
       \Email{edwardchoi@kaist.ac.kr} \\ 
       \Name{Byungeun Ahn$^{1}$}
       \Email{ben@aitrics.com} \\ 
       \Name{Joohyung Lee$^{1}$\thanks{Corresponding author} \Email{chris@aitrics.com}\\}
       \addr$^{1}$AITRICS Inc.\\
        \addr$^{2}$Korea Advanced Institute of Science and Technology (KAIST)\\
        \addr$^{3}$Sungkyunkwan University (SKKU)} 

\begin{document}

\maketitle

\begin{abstract}
    Electronic Health Record (EHR) provides abundant information through various modalities. However, learning multi-modal EHR is currently facing two major challenges, namely, 1) data embedding and 2) cases with missing modality. A lack of shared embedding function across modalities can discard the temporal relationship between different EHR modalities. On the other hand, most EHR studies are limited to relying only on EHR Times-series, and therefore, missing modality in EHR has not been well-explored. Therefore, in this study, we introduce a Unified Multi-modal Set Embedding (UMSE) and Modality-Aware Attention (MAA) with Skip Bottleneck (SB). UMSE treats all EHR modalities without a separate imputation module or error-prone carry-forward, whereas MAA with SB learns missing modal EHR with effective modality-aware attention. Our model outperforms other baseline models in mortality, vasopressor need, and intubation need prediction with the MIMIC-IV dataset.
    
\end{abstract}
% So far, no single unified embedding method for multi-modal EHR data has been introduced, to the best of our knowledge.

% Usually, using EHR Time-series was complicated by data irregularity/asynchrony. For modalities other than EHR Time-series, time information was usually discarded.

\input{Sections/2_Introduction}

\input{Sections/3_Generalizable_Insights}
\input{Sections/4_Related_Works}
\input{Sections/5_Methods}
\input{Sections/6_Experiments}

\input{Sections/7_Results}

\input{Sections/8_Discussion}

\bibliography{ref}
\input{Sections/9_Appendix}

% \appendix
% \section*{Appendix A.}

% Some more details about those methods, so we can actually reproduce
% them.  After the blind review period, you could link to a repository
% for the code also.  \emph{MLHC values both rigorous evaluation as well
%   as reproduciblity.}

\end{document}

%% file: Sections/2_Introduction.tex
\section{Introduction}
    \label{sec:introduction}
    % EHR 데이터 중요. EHR과 ml 엮음. ehr의 clinical role
    Recently, electronic health record (EHR) emerges as a promising source of patient information. Utilizing its rich information, deep learning is making significant progress in various clinical regimes, especially in event prediction, e.g., mortality, sepsis, acute kidney injury, as well as the need for vasopressor administration, intubation, and ICU transfer (\cite{sung2021event, wanyan2021contrastive}). Clinically, the early prediction of clinical events enables clinicians to effectively prioritize high-risk patients, allocate resources efficiently, and make prompt interventions (\cite{choi2022advantage}). Nevertheless, two problems in EHR are hindering many promising deep learning algorithms to be readily transferred to learn EHR: missing modality and irregular/asynchronous sampling.
    
    % EHR은 여러 데이터 종류 포함하는데, 대부분 타임데이터 사용. 왜냐? 미싱 ㅡㅡ
    EHR encompasses a wide range of modalities, including not only EHR time series but also medical images (e.g., X-ray images), text (e.g., clinical notes, chief complaints), and demographics, all of which hold the potential for enhancing the predictive performance of clinical event (\cite{lee2022self, hayat2022medfuse}). Among various modalities in EHR, time-series data is most frequently used for clinical event prediction, and many reported EHR studies solely rely on time-series EHR data (i.e., vital signs and laboratory test results) (\cite{wanyan2021contrastive, choi2022advantage, sung2021event, kim2019deep, che2018recurrent, shukla2019interpolation,tipirneni2022self}). Though the predictive performance from EHR time series can be improved by supplementing other modalities \cite{lee2022self}, multi-modal learning has not been widely explored in EHR learning.
    
    One of the major challenges in multi-modal EHR learning is the missing modality. Specifically, in practice, not all data modalities are consistently available for patients. Frequent modality missing in EHR data impedes the use of multi-modal fusion models to fuse a wide range of EHR data. Moreover, a weak inter-modality relationship and varying dimensionalities of different EHR modalities even complicate learning multi-modal EHR with missing modalities.

    % In prior studies, some researchers tackled the missing modality issue by employing generative methods to create representations for missing modalities using existing ones (\cite{Ma2021SMILML, Vasco2020MHVAEAH}). However, these approaches are unsuitable for medical multimodal data due to its heterogeneous nature (e.g., time-series pulse data cannot generate X-ray images).
    Various studies have addressed the missing modality problem. For example, many studies have approached the missing modality problem with generative methods (\cite{Ma2021SMILML, Vasco2020MHVAEAH}), but the generative method is unsuitable for learning multi-modal EHR due to the excessive heterogeneity of modalities of EHR (e.g., time-series pulse data cannot generate X-ray images). \cite{ma2022multimodal} introduced self-attention masking for missing text modality but reported performance degradation when trained with missing-modal data. Moreover, because they modeled the relationship between every possible token pair regardless of the token modality, the computational cost increases much with the increasing number of modalities ($O(n^{2})$), which can thus be less scalable for multi-modal fusion. \cite{hayat2022medfuse} utilized basic LSTM structure to late-fuse the X-ray imaging to the time-series EHR in mortality and phenotype prediction. LSTM can handle the missing modality problem, for LSTM can function with variable input length.
    
    % Previous studies primarily targeted bimodal settings, whereas \cite{soenksen2022integrated} tackled missing modalities in a four-modality context using basic concatenation for fusion. This limited the examination of diverse EHR multimodal data types. Considering previous research, a model that effectively handles missing modalities and deeply explores multimodal interactions is necessary for improving clinical event predictions.
    
    \input{Sections/Figure1.tex}
    
    % irregular/async: 소개 -> 여러 방법들이 있음 (마지막에 set 방식)
    % 사실 irregular 하고 asnyc 한건 time-series 만이 아니고 static feature (demo, etc)를 제외한 모든 놈들이 그럼. time-series 말고 다른 모달리티의 시간 정보를 활용한 케이스 -> 우리는 unified.
    Other than the missing modality, a lack of shared embedding functions across EHR modalities can be problematic since a unified embedding method can model the temporal relationship between different modalities. Reported EHR embedding studies usually address irregularity/asynchrony in EHR time-series data only \cite{horn2020set,choi2022advantage,tipirneni2022self}. However, other EHR modalities, e.g., medical images, clinical notes, and laboratory test results, are recorded at irregular time intervals as well, depending on factors such as clinical protocols, patient conditions, and healthcare settings (\cite{che2018recurrent,shukla2019interpolation,tipirneni2022self,zhang2020time}). In fact, \cite{zhang2020time} have shown that learning clinical notes without considering their occurrence time information can lead to misclassification. Yet, no single unified embedding function method for different modalities has been introduced. Therefore, in this study, we propose a unified set embedding that addresses the irregularity/asynchrony of all modalities without a separate imputation module. In summary, our contributions are:

    \begin{itemize}
    \item We propose a Unified Multi-modal Set Embedding (UMSE) as an efficient embedding method for Multi-modal EHR. UMSE views all modalities in the same line and provides a unified method to 1) solve irregular/asynchronous problems of all EHR modalities, 2) utilize the time information of all EHR modalities, and 3) model the temporal relationship between different modalities by sharing the time embedding function across different modalities.

    \item We suggest Modality-Aware Attention (MAA) and Skip-Bottleneck fusion (SB) to effectively learn multi-modal EHR with modality missing. MAA assigns distinct attention to each modality compared to the averaging method from Multi-modal Bottleneck Transformer (MBT) \cite{Nagrani2021AttentionBF} whereas SB enables MBT to learn with missing modality.
    
    % By using bottleneck tokens for multi-modal fusion, we aim to reduce the computation burden when increasing the modality, and thus more scalable.
    
    % We present ㅇㅇㅇ which effectively solves two prevalent problems in learning EHR, namely, missing modality and irregular/asynchronous sampling. First, ㅇㅇㅇ proposes skip-bottleneck and modality-aware attention for the missing-modality problem in EHR. We use bottleneck fusion due to 1) its performance superiority (\ref{tab2}) and 2) better scalability, which comes from the less computation burden than Multi-modal Transformer \cite{ma2022multimodal}.
    
    % MBTI, a multi-modal model that integrates time-series numeric, X-ray image, and clinical text data for patient outcome prediction. MBTI effectively handles EHR multimodal data with missing modal cases by simple yet effective fusion and optimization methods. Moreover, the model accommodates missing modalities during both training and inference, and we demonstrate its effectiveness by comparing its clinical event prediction performance against existing EHR models that handle missing modalities.
    
    \item We provide extensive analysis of each component of our proposed model to show the effects of different design choices. Here, we discuss 1) the application of UMSE on different modalities, 2) MAA comparison, 3) the effects of pretraining (Appendix \ref{tab8}) and 4) modality combinations.
    
    \item In this paper, we extensively experimented with three different clinical tasks: mortality, vasopressor need, and intubation need predictions with publicly open-access real-world large dataset Medical Information Mart for Intensive Care (MIMIC-IV), MIMIC Chest X-ray (MIMIC-CXR) and MIMIC Emergency Department (MIMIC-ED) (\cite{johnson2020mimic, johnson2019mimic, johnsonmimiced}). With a real-time time embedding (Appendix \ref{realtimeresults}) for an online monitoring scenario using maximum 1440 hours of each subject, we estimated the performance in a practical setting.

    \item Lastly, considering the lack of benchmark in the field of EHR multi-modal fusion with missing modality while handling irregular sampling, we release our code to ensure the reproducibility and applicability of our approach: \url{https://github.com/anonymous/Trimodal_EHR.}
    
    \end{itemize}  

    % These previous works proposed concatenation strategy to fuse data from different modality (\cite{lee2022self, hayat2022medfuse}) or simply replicated irregularly sampled modality data into dense form (\cite{Suresh2017ClinicalIP})
    
    % In this paper, we explore the importance of early prediction of mortality, vasopressor usage, and intubation usage in the context of clinical decision-making and patient outcomes. The timely and accurate prediction of patient critical events can provide healthcare professionals with valuable information to guide their interventions, enabling them to initiate appropriate treatment strategies and allocate resources more efficiently. Early identification of patients at high risk for mortality or requiring advanced life support measures, such as vasopressor administration or intubation, allows for proactive management, which may help reduce the severity of complications, optimize patient care, and ultimately improve overall patient outcomes. By developing and implementing advanced predictive models based on multimodal electronic health record data, clinicians can enhance their ability to recognize and respond to these critical events, fostering a more personalized and effective approach to patient care.

%% file: Sections/Figure1.tex
\begin{figure}[t]
\centering
\includegraphics[width=\textwidth]{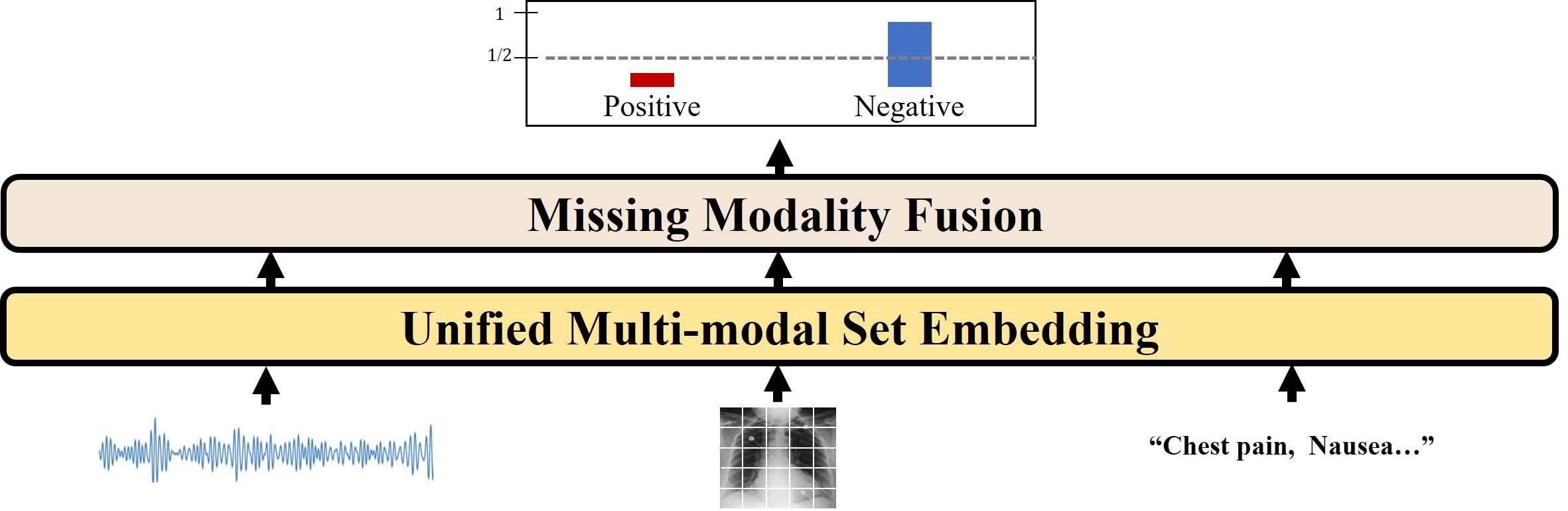}
\caption{Overview of our proposed method consisting of 1) \textbf{Unified Multi-modal Set Embedding (UMSE)} which embeds accurate time and feature type information to all modality data and 2) \textbf{Missing Modality Fusion}.}
\label{fig1}
\end{figure}

%% file: Sections/3_Generalizable_Insights.tex
\subsection*{Generalizable Insights about Machine Learning in the Context of Healthcare}
    \label{sec:generalizable_insights}
    Learning multi-modal EHR usually involves two problems: 1) irregular/asynchronous data and 2) missing modality. Though many studies have proposed to solve irregular/asynchronous data problems, they are limited to apply their method for time-series EHR. However, \cite{zhang2020time} have demonstrated the need for the time information on EHR Text as well, and a lack of shared embedding method across modalities can discard the important temporal relationship between different EHR modalities. On the other hand, \cite{lee2022self} have shown that multi-modal bottleneck fusion outperforms other regressors such as vanilla Transformer in prediction tasks using EHR. However, bottleneck fusion (BF) has two drawbacks to be readily applied to learn multi-modal EHR: 1) BF does not allow missing modality, 2) BF neglects different importance of modalities since it computes the final logit by averaging logits from each modality. To tackle the aforementioned problem, we suggest a Unified Multi-modal Set Embedding (UMSE), Skip Bottleneck (SB), and Modality-Aware Attention (MAA). First, UMSE effectively handles irregular/asynchronous data problems of all modalities, and it can encourage modeling the temporal relationship between different modalities by sharing the time embedding function across all modalities. Second, SB empowers bottleneck fusion to handle missing modalities. Lastly, MAA employs modality-aware attention scores to compute the final logit.
    
    % and 2) BF disregards varying modality importance by averaging logits, which can be problematic in EHR data where importance of each modality can be varying according to prediction task. This is different from conventional multimodal models where all modalities such as visual, audio, and text data are similary important.

%% file: Sections/4_Related_Works.tex
\section{Related Works}
    \label{sec:related_works}
    % In this section, we provide an overview of related works on clinical event prediction considering the temporal embedding of multimodal EHR data while handling missing modalities. The growing interest in developing advanced machine learning models to predict clinical outcomes from EHRs has led researchers to explore various techniques. 

    \subsection{Learning Multi-Modal EHR for Event Prediction}
    Efforts to obtain a more comprehensive understanding of patient patterns for accurate clinical event prediction have led numerous researchers to employ multi-modal EHR datasets. Many have reported clinical event prediction methods using various EHR data combinations such as time-series EHR combined with clinical text (\cite{lee2022self, wang2022multi, Suresh2017ClinicalIP, qin2021improving, lyu2022multimodal}), medical codes (e.g., procedure code, diagnosis code) alongside EHR text (\cite{qiao2019mnn}). \cite{choi2022advantage} have incorporated time-series vital signs with wearable device-based heart signal data for clinical event prediction. Meanwhile, \cite{vale2020multisurv} have employed patient genetics, clinical, and histopathology slide images for long-term pan-cancer survival prediction.
    
    \subsection{Multi-Modal Fusion}
    Numerous research methodologies have been investigated to obtain extensive knowledge from multi-modal data. To learn both tri-modal and bi-modal interactions, \cite{Zadeh2017TensorFN} employed the Tensor Fusion method. Several studies have utilized the Transformer architecture (\cite{Vaswani2017AttentionIA}) for multi-modal fusion; \cite{kim2021vilt} adopted Transformer structure to integrate multi-modal data, without requiring encoders dedicated to each modality; \cite{Tsai2019MultimodalTF} used cross-modal attention module to learn varied bi-modal interactions of the multi-modal data; \cite{akbari2021vatt, Alayrac2020SelfSupervisedMV} applied self-supervision techniques to train modality-specific encoders to project their modality representations to a common space dimension for improved downstream performance in learning video/audio/text; and \cite{Nagrani2021AttentionBF} employed bottleneck learnable tokens to model inter-modality interaction with the Transformer architecture without much-increasing computation burden.
    
    \subsection{Missing Modality}
    In practice, it is often the case that not all modalities are available for every patient, leading to the issue of "missing modality". Intuitively, many researchers have explored methods for handling missing modalities by generating representative vectors. \cite{Ma2021SMILML, Vasco2020MHVAEAH} learned to generate representative vectors of missing modalities. \cite{Poklukar2022GeometricMC} trained encoders to make  representations of all missing modality combinations similar to the representation of the full modality combination. \cite{ma2022multimodal} simply employed a masking method for self-attention to address missing modality and suggested layer fusion with multi-task learning (multi-token) to enhance model robustness to missing modality. \cite{hayat2022medfuse} used LSTM to fuse modality-wise representations with possible modality-missing. \cite{vale2020multisurv} filled missing feature values with median substitution.

    % majority of machine learning multimodal fusion approaches concentrate on synchronous scenarios, where all modality data are regularly sampled and fully paired (\cite{lee2022self, kim2021vilt, akbari2021vatt}).
    % A few simply overcome these problems with carry-forward (\cite{lee2022self,Suresh2017ClinicalIP}), zero-padding (\cite{soenksen2022integrated}) or simply used LSTM recurrent structure to handle missing modality (\cite{hayat2022medfuse}).
    
    \subsection{Data Embedding in EHR}
    Various embedding strategies have been reported to incorporate the occurrence time of EHR data. \cite{che2018recurrent} utilize each feature's missing period as their temporal information. \cite{choi2022advantage,lee2022self,hayat2022medfuse} used 1 or 2 hourly sampling method to discretize the temporal axis of time series EHR data. \cite{horn2020set} suggested a set encoding method to train the model with irregularly sampled time-series EHR data without carry-forward or separate generative module. \cite{tipirneni2022self} employed learnable set embeddings in predicting mortality from EHR time-series data. \cite{zhang2020time} devised Flexible Time-aware Long Short Term Memory (FT-LSTM) in order to use both time information and hierarchical information of clinical text. 

    % intro의 글과 중복됨

    % 1. ehr multi-modal learning for event prediction
    % 2. multi-modal fusion
    % 3. missing modality
    % 4. vslt embedding: triplet embedding and other temporal embedding method

%% file: Sections/5_Methods.tex
\section{Methods}
    \label{sec:methods}
    % Our method의 주요 내용 짧게 설명. 앞으로 있을 논문의 모든 모델들에서 Classifier Layers가 똑같은점 설명
    In this section, we describe our method to effectively learn multi-modal EHR data with modality missing. Our model aims to tackle two problems in learning multi-modal EHR, namely, 1) EHR data embedding, and 2) modality fusion with missing modality. As depicted in Figure~\ref{fig1}, our Unified Multi-modal Set Embedding (UMSE) and Missing Modality Fusion module (MMF) address these problems. Specifically, UMSE allows using irregular/asynchronous multi-modal EHR data with neither a separate imputing module nor loss of time information. Moreover, UMSE can model inter-modality temporal relationships by sharing the time embedding function across modalities. On the other hand, MMF enables the model to learn multi-modal data with possible modality-missing through Skip Bottleneck (SB). Moreover, through Modality-Aware Attention (MAA), MMF assigns different attention scores for each modality to enhance the predictive performance. All three prediction tasks (i.e. mortality, vasopressor need, and intubation need prediction) are binary classification tasks predicting whether the event would occur within 12 hours.
    \input{Sections/Figure2.tex}
    \subsection{Unified Multi-modal Set Embedding}
    Our Unified Multi-modal Set Embedding (UMSE) aims to tackle 1) the irregular and asynchronous nature of EHR multi-modal data, and 2) inter-modality temporal relationships. In practice, both irregularity and asynchrony occur in time-series EHR, whereas only irregularity exists in other EHR modalities such as EHR Image. Traditionally, error-prone carry-forward or separate imputation modules have been widely employed for the time-series EHR, and for other EHR modalities, time information is usually discarded.
    
    \vspace{3mm}
    \textbf{Definition 1} (Irregularity). \textit{We consider an arbitrary EHR feature $B$ occurs N times, i.e. $B := \{(\,S_1, t_1)\, , ..., (\,S_N, t_N)\, \}$, where $t_n$ denotes the occurrence time of the feature $D$ with its value $S_n$ and $t_i < t_{i+1}$. A feature $B$ is irregularly sampled if there exists at least one $t_i$ such that $t_{i+1}-t_{i} \not= t_{i}-t_{i-1}$.}

    % \textbf{Definition 2} (Asynchrony). \textit{A D-dimensional EHR feature, i.e., $\{(\,S_1, t_1)\, , ..., (\,S_N, t_N)\ \}$ is asynchronous if there exists at least one occurrence time $t_i$ at which at least one element is missing, i.e., $|S_i|\not=D$.}  
    \textbf{Definition 2} (Asynchrony). \textit{A D-dimensional EHR feature $B$ occurs N times, i.e., $B := \{b_1, , ..., b_N \}$. A feature $B$ is asynchronous if there exists at least one $b_i$ at which at least one element is missing, i.e., $|b_i|\not=D$.}

    \textbf{Definition 3} (Multi-modal EHR). \textit{We denote a multi-modal EHR data of $i^{th}$ subject as a set $S_i$ of N:= $|S_i|$ observations $s_i$ where $S_i$ := $\{s_1, ..., s_N\}$. We treat each observation $s_i$ as a triplet $(\,v_i, t_i, FT_i)\,$, consisting of an observed value $v_i\in\mathcal{R}^{M_{FT_i}}$, observation time $t_i\in\mathcal{R}$, observed feature type indicator $FT_i\in\{1, ..., D\}$, where $D$ represents the dimensionality of the whole multi-modal EHR including not only EHR image, EHR text, but also each feature in EHR Time-series, e.g., Hematocrit, Lactate, etc. $M_{FT_i}=1$, $224\times224$ when $FT_i$ is EHR Time-series, EHR image, respectively. For EHR text, $v_i$ is not numeric but text string.}

    \vspace{3mm}
    Inspired by \cite{horn2020set,tipirneni2022self}, our UMSE rephrases the problem of encoding multi-modal EHR into the problem of encoding a set of observations as described in the above definition. To this end, our UMSE consists of three embedding functions as illustrated in Figure~\ref{fig2}: value embedder, time embedder, and feature type embedder, which are denoted as $Embedder_{Value}$, $Embedder_{Time}$, $Embedder_{Feature Type}$, respectively. These three embedders encode each element of the observed triplet and add them up. The output of the UMSE is then concatenated with the outputs from other observations.
    
    We use a modality-specific encoder for the value embedder. Specifically, we use pre-trained frozen Swin-Transformer \cite{liu2021swin} followed by a linear projection for the EHR image. For EHR text, we used a BERT tokenizer and pre-trained \& frozen BioBERT \cite{lee2020biobert} followed by a linear projection. For EHR Time-series, we used a simple linear projection with nonlinearity. Please refer to Appendix \ref{pretrainedimage}, \ref{pretrainedtext} for more details.
    
    The dimension of the input/output of the value embedder varies with the modality, i.e., EHR Time-series, EHR image, and EHR text. Specifically, $Embedder_{Value}: \mathcal{R}\rightarrow\mathcal{R}^{256}$, for EHR Time-series, $Embedder_{Value}: \mathcal{R}^{224\times224}\rightarrow\mathcal{R}^{49\times256}$ for EHR image,  $Embedder_{Value}: \mathcal{S}\rightarrow\mathcal{R}^{128\times256}$, for EHR text ($\mathcal{S}$ denotes string).

    Time embedder and feature type embedder are shared across all modalities to model inter-modality temporal relationships. Inspired by \cite{tipirneni2022self}, we use a look-up table and simple linear projection with nonlinearity for the feature type embedder and time embedder, respectively.

    \input{Sections/Figure3.tex}
    \subsection{Missing Modality Fusion}
    Our Missing Modality Fusion module (MMF) is depicted in Figure~\ref{fig1} and in Figure~\ref{fig3} models the intra-modality and inter-modality interaction for three different clinical prediction tasks, i.e., mortality, vasopressor need, and intubation need prediction. Our MMF consists of two modules, i.e., Skip Bottleneck (SB) and Modality-Aware Attention (MAA). SB enables the Multimodal Bottleneck Transformer (MBT) to handle data with missing modality whereas MAA provides modality-wise attention to consider the logit from different modality-transformer differently. All of our MAA processes are applied on top of the MBT with SB for cases involving missing modalities.

    \subsubsection{Revisiting Bottleneck Fusion}
    \cite{Nagrani2021AttentionBF} proposes the MBT (Multimodal Bottleneck Transformer) architecture, which effectively reduces the computational costs of transformer models. MBT enables modality interaction exclusively via fusion bottleneck tokens shared among modality-specific transformer layers (Equation \ref{eq:3}). This structure encourages intra-modality interaction while managing inter-modality interaction through a narrow bottleneck, which may be advantageous for EHR multimodal data due to its inherent heterogeneity and rather weak correlations. In a multi-modal transformer, \cite{ma2022multimodal} simply used the attention masking method to handle missing modality during inference. The masked attention excludes missing modality in self-attention softmax calculation which prevents unnecessary interaction between observed modality data and unobserved missing modality data.
    
    \subsection{Skip Bottleneck Transformer}
    \cite{lee2022self} have shown that Multimodal Bottleneck Transformer (MBT) \cite{Nagrani2021AttentionBF} outperforms Multimodal Transformer (MT) \cite{ma2022multimodal} and other regressors in multi-modal EHR learning. However, in practice, a single mini-batch comprises subjects with various combinations of modalities, and MBT cannot process a mini-batch with varying modalities. To supplement MBT for missing modality, we propose Skip Bottleneck (SB).
    
    SB consists of two simple processes; 1) we feed random numbers (e.g. zero-vectors) to the Transformer of the missing modalities; 2) we `skip' the bottleneck tokens from the missing modality for fusion (Equation \ref{eq:4}) (Figure \ref{fig3}).
    
    % \begin{equation}\label{eq:3}
    % |z_{cls^m}^{l+1}||\hat{z}_{fusion}^{l+1}| = Transformer(|z_{cls}^{l}||z_{fusion}^{l}|;\theta)
    % \end{equation}
    \begin{equation}\label{eq:3}
    |z_{i}^{l+1}||\hat{z}_{fsn_i}^{l+1}| = Transformer(|z_{i}^{l}||z_{fsn}^{l}|;\theta_i)
    \end{equation}
    
    % \begin{equation}\label{eq:4}
    %     |z_{fusion}^{l+1}| = Average(\{z_{fusion}^{1:\hat{M}}=(z_{fusion}^{1}, ..., z_{fusion}^{\hat{M}})\})
    % \end{equation}
    \begin{equation}\label{eq:4}
    \begin{aligned}
        \ \ \ \ \ \ \ \ \ z_{fsn}^{l+1} &= \frac{1}{1+\mathbbm{1}_{Image}+\mathbbm{1}_{Text}}(\hat{z}_{fsn_{Time-Series}}^{l+1}+\mathbbm{1}_{Image}\hat{z}_{fsn_{Image}}^{l+1}+\mathbbm{1}_{Text}\hat{z}_{fsn_{Text}}^{l+1}),\\
        \mathbbm{1}_{i} &= 
        \begin{cases}
        1,\hspace{0.5cm} \text{if modality i is present}\\
        0,\hspace{0.5cm} \text{if modality i is NOT present}
        \end{cases}        
    \end{aligned}
    \end{equation}

    $z_i^{l}$ refers to the token at the Transformer layer $l$ of the observed modality $i$, whereas $\hat{z}_{fsn_i}$ and $z_{fsn}$ refers to the bottleneck fusion token before and after averaging, respectively. As illustrated in Figure~\ref{fig3} and Equation \ref{eq:4}, the temporary bottleneck token $\hat{z}_{fsn_i}$ from EHR Time-series Transformer is never skipped since subjects in multi-modal EHR always possess Time-series data.
    
    % \begin{equation}\label{eq:5}
    %     |z_{fusion}^{l+1}| = \hat{z}_{fusion}^l
    % \end{equation}

    \subsection{Modality-Aware Attention Decision Making}
    The original MBT (\cite{Nagrani2021AttentionBF}) averages the pre-softmax logits of $CLS_M$ tokens before feeding them to the shared classification layer. This approach inherits the assumption that all modalities are 'equally' significant in the decision-making process for visual-audio tasks. However, in EHR multi-modal learning, time-series data is often regarded to be more significant than other modalities. Consequently, we added two more designs to the traditional Average Attention (AA) to experiment with different modality attention: 1) Time-Series Attention (TSA), and 2) Clinical-Task-Aware Attention (CTAA) as described in (Figure \ref{fig3}). Specifically, TSA places the [CLS] token solely for the Time-series Transformer, while CTAA employs learnable scalars with temperature $\tau$. As a result, CTAA creates a modality-wise attention score through the softmax function and determines which modality's logit should be prioritized (Equation \ref{eq:5-1}). Note that all TSA, CTAA and AA are built on top MBT with SB with different MAA strategies.

    % P_{attention} = Softmax([W_{TimeSeries}, W_{Text}, W_{Image}] \times \tau) 
    
    \begin{equation}\label{eq:5-1}
    Attention_{m} = \frac{exp(w_{m}/\tau)}{\sum_{j}exp(w_{j}/\tau)}
    \end{equation}

    where $w$ are three learnable logits of softmax and $\tau$ is the temperature to sharpen the softmax function to get a pre-sigmoid-logit before binary cross-entropy calculation. Note that $m$ is a modality indicator.
      

%% file: Sections/Figure2.tex
\begin{figure}[t]
\centering
% \vspace{-12}
\includegraphics[width=\textwidth]{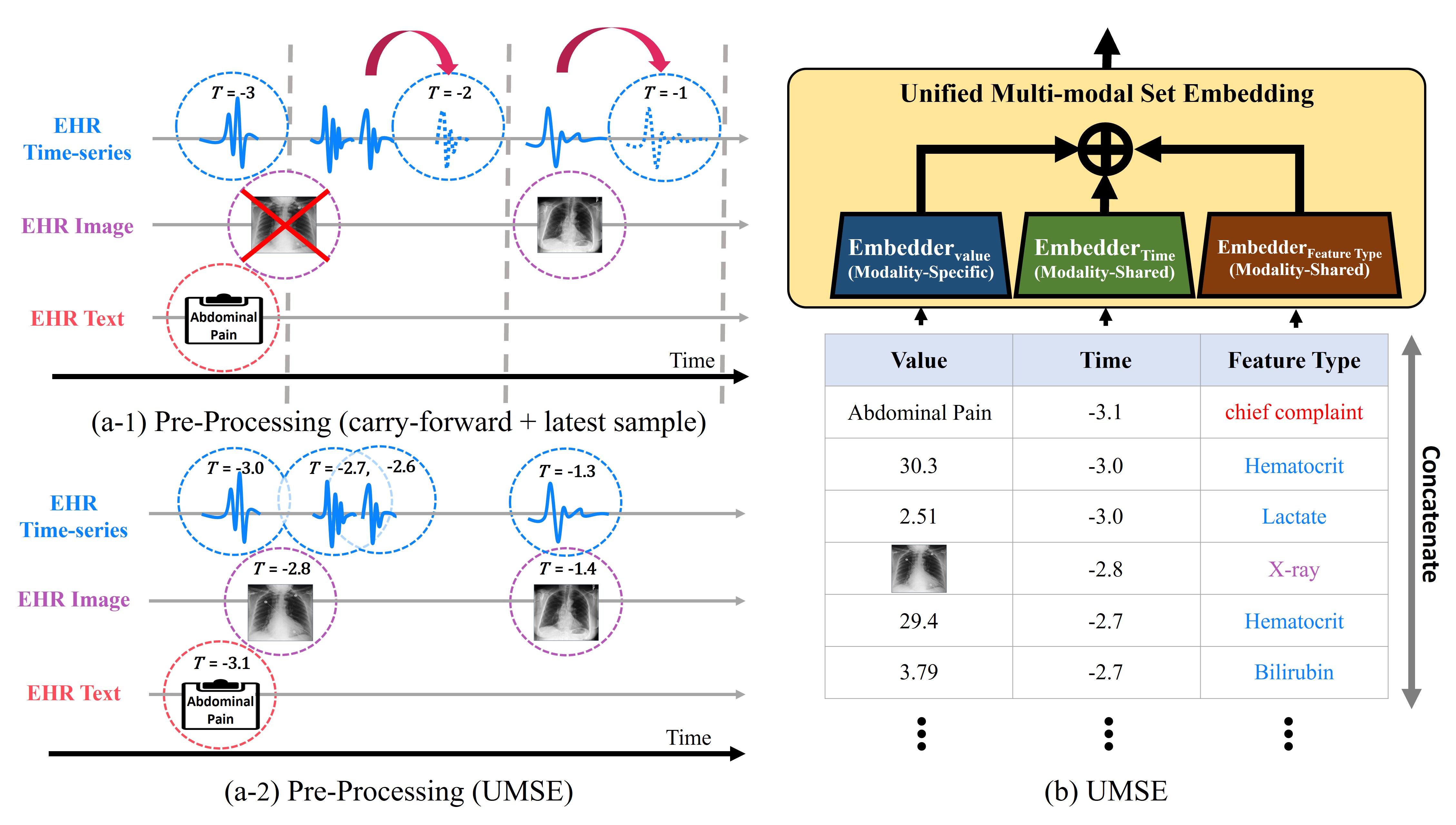}
\caption{(a-1) Overview of traditional preprocessing strategy for EHR multi-modal data: carry-forward for regular time-grid and latest sampling for non-Time-series modality (\cite{lee2022self,hayat2022medfuse}). (a-2, b) Our Unified Multi-modal Set Embedding (UMSE).}
\label{fig2}
% \vspace{-12}
\end{figure}
 % which accurately embed time and feature information to multimodal data.
% time-series data and use only the most recent image and text data without considering temporal information

%% file: Sections/Figure3.tex
\begin{figure}[t]
% \centering
% \vspace{-12}
\includegraphics[width=\textwidth]{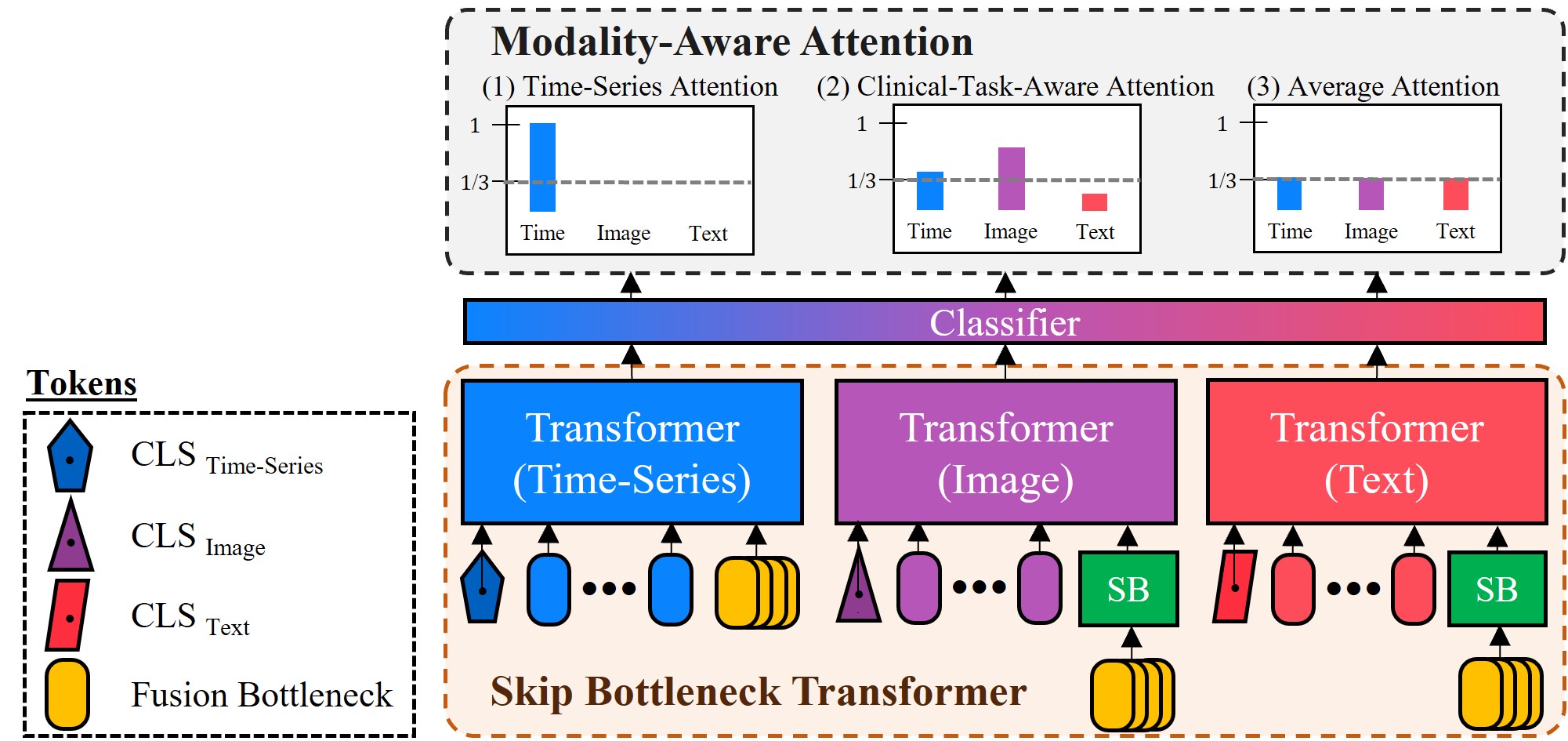}
\caption{Overview of our Missing Modality Fusion module (MMF) consisting of Skip Bottleneck (SB) Transformer with three different modality-aware attention schemes: (a) Time-Series Attention (TSA), which focuses on Time-series modality only, (b) Clinical-Task-Aware Attention, which dynamically attends modalities depending on target task and observing modalities, (c) Average Attention (AA), which considers all modalities equally. Note that TSA, CTAA, and AA are built on top of MBT with SB. Note that AA is simply MBT with SB.}
\label{fig3}
% \vspace{-12}
\end{figure}

%% file: Sections/6_Experiments.tex
\section{Experiments}
    \label{sec:experiment}

    \subsection{Dataset}
    % In this paper, we utilized three datasets: Medical Information Mart for Intensive Care (MIMIC-IV), MIMIC Chest X-ray (MIMIC-CXR) and MIMIC Emergency Department (MIMIC-ED) (\cite{johnson2020mimic, johnson2019mimic, johnsonmimiced}). Since all three datasets share same patients, we merged the datasets so we could obtain numeric data such as vital-sign, laboratory results, and demographic data from MIMIC-IV, X-ray imaging data from MIMIC-CXR and Chief-Complaint text data from MIMIC-ED for each patient. For additional text data experiment, we employed the newly-released MIMIC-IV-Note dataset, extracting clinical note (\cite{johnsonmimicnote}). We derived three cohorts from the combined MIMIC data: clinical event like mortality and clinical intevention such as vasopressor need, or intubation need. The details of data is outlined below. 
    In this study, we use three EHR datasets: MIMIC-IV\footnote{\url{https://physionet.org/content/mimiciv/1.0/}}, MIMIC-CXR\footnote{\url{https://physionet.org/content/mimic-cxr/2.0.0/}}, and MIMIC-ED\footnote{\url{https://physionet.org/content/mimic-iv-ed/2.2/}} (\cite{johnson2020mimic, johnson2019mimic, johnsonmimiced}). As they share the same patients, we merged them to obtain per-patient information of vital signs, lab results, demographics from MIMIC-IV, X-ray images from MIMIC-CXR, and chief-complaint text from MIMIC-ED. Additionally, we used the MIMIC-IV-Note\footnote{\url{https://physionet.org/content/mimic-iv-note/2.2/}} dataset for clinical notes (\cite{johnsonmimicnote}) and compared the performance when it replaces the chief-complaint in Appendix \ref{clinicalnoteresult} with a data table Appendix \ref{ehrtextdata}. Unless otherwise specified, EHR text data refers to chief-complaint from MIMIC-ED throughout this study. If chief-complaint text data exists, it means that the patient has visited ED before ICU admission. Details are provided below.
    \input{Sections/Table1_data}

    \subsubsection{Data Preprocessing}
    \label{sec:datapreprocess}
    \begin{itemize}  
    \item \textbf{EHR Times-series and Demographic data}: 
    For each ICU patient, we collected EHR numeric data from MIMIC-IV, ranging from a minimum of 3 hours to a maximum of 1440 hours (60 days). Numeric data comprises demographic features (age and gender), as well as time-series data, i.e., vital signs and lab-test results. Vital sign includes six features: heart rate, respiration rate, diastolic and systolic blood pressure, temperature, and pulse oximetry. The laboratory result data, i.e. lab-test, encompasses ten features: Hematocrit, Platelet, Bilirubin, etc, following \cite{sung2021event} (See Appendix \ref{ehrtimedata}). In total, there are 18 numeric data features. We exclude patients without (or less than) 5 vital-sign features during the entire ICU hospitalization period. We applied min-max normalization using our training set. More detailed information regarding our EHR time-series data is provided in Appendix \ref{ehrtimedata}.
    
    \item \textbf{EHR Text data}: We extract chief-complaint text from MIMIC-ED and admission-related text (Chief Complaint, Medication on Admission, Past Medical History) from MIMIC-IV-Note. We described more detailed information about EHR text data preprocessing steps in Appendix \ref{ehrtextdata}.
    
    \item \textbf{X-ray Image data}: We preprocess MIMIC-CXR X-ray image by removing the black margin area and excluding images with aspect ratios bigger or less than 1.3 or 0.7 respectively. More detailed image preprocessing information, pre-training strategy, and image augmentation method are described in Appendix \ref{ehrimagedata}.
    
    \end{itemize} 
    
    For training, we randomly selected a time window ranging from 3 to 24 hours to predict the occurrence of clinical events within the next 12 hours. We excluded any time windows with no EHR Time-series data within the most recent hour interval, which is the latest 1-hour within the training time windows, for both training and inference. During training, we extracted positive and negative windows with equal ratios using a batch sampler as described in Appendix \ref{datasampler}. For inference, we randomly selected and fixed 5 positive and 5 negative periods per patient during ICU hospitalization period.
    
    \subsubsection{Data Split}
    We randomly selected 80\%, 10\%, and 10\% of patients for training, validation, and test set. For each patient, we extracted EHR Time-series data with EHR Text data and EHR image data from MIMIC-ED and MIMIC-CXR with date time information indicating when the text or image was captured. Table \ref{tab1} illustrates that not all modalities are paired for each sample, and provides missing rate information.

    \subsection{Clinical Objectives}
    \label{sec:taskdataset}
     We extracted three tasks-related information, i.e., mortality, vasopressor need, and intubation need predictions with the following statistics.
     
    \begin{itemize}  
    \item \textbf{Mortality prediction within 12 hours}: As depicted in Table \ref{tab1}-(a), we utilized 42,813 ICU cases, comprising 4,329 positive cases with defined mortality onset times.
    % Cases with uncertain death times were excluded.

    \item \textbf{Vasopressor need prediction within 12 hours}: As depicted in Table \ref{tab1}-(b), we utilized 42,572 ICU cases, including 11,696 positive cases labeled with vasopressor initiation times. Labels were assigned when Norepinephrine, Dopamine, Dobutamine, or Epinephrine was administered. The Appendix \ref{appendixtasks} contains item number details.
    
    \item \textbf{Intubation need prediction within 12 hours}: As outlined in Table \ref{tab1}-(c), we extracted 42,532 ICU cases from which 16,831 cases were labeled positive with intubation start times. We focused on 7 intubation types among diverse MIMIC-IV chart events. Specific intubation types and item numbers are detailed in Appendix \ref{appendixtasks}.
    \end{itemize}  
    
    %%%
    %%% Model Experiments
    %%%
    \subsection{Baseline Models}
    We compare the performance of our model using test set AUPRC and AUROC with previously reported EHR multi-modal algorithms. Since our prediction tasks are highly imbalanced, we primarily focus on AUPRC and considered AUROC secondarily. 
    % All reported performances are the average of 3 seeds with their standard deviation value. 
    To ensure a fair comparison, all classification layers in this paper consist of a 2-layer Multi-Layer Perceptron (MLP) with Layer Normalization (LN) and ReLU non-linearity between the two linear layers. We process EHR static features, i.e., age and gender, through one linear projection with ReLU nonlinearity; we concatenate it to the output of all fusion algorithms (e.g., Transformer). All transformer fusion networks have 6 layers with 256 feature dimensions. We conducted a learning rate sweep ranging from $10^{-6}$ to $10^{-4}$. All models are trained with AdamW optimizer, 50 epochs with a 3-seed averaging. We compare our model performance against the following algorithms: 
    \begin{itemize}
    
    % \item \textbf{Unimodal Models}(Numeric): We measured the performance of unimodal models that utilize only time-series data from MIMIC-IV. The compared unimodal models are 1) LSTM (\cite{sak2014long}), 2) GRU-D (\cite{che2018recurrent}), 3) Transformer with hourly interval carry-forward numeric data (\cite{Vaswani2017AttentionIA}), 4) Transformer with initial triplet embedding numeric data (\cite{tipirneni2022self}).
    
    \item \textbf{HAIM}: HAIM (\cite{soenksen2022integrated}) used varying pre-trained modality-specific encoders for each modality. The encoder outputs are then concatenated before linear classification layers. For missing modalities, a zero-padding strategy is employed.
    % In this paper, we experimented the network with numeric, image and text multimodal data.
    
    \item \textbf{MNRIFN}: Proposed by \cite{wang2022multi}, this method only accommodates sequential multi-modal data; we reproduced using our time-series and text data. 
    % We implement a multi-view fusion approach that derives non/-redundant information from both text and clinical time-series data.

    \item \textbf{Medfuse}: \cite{hayat2022medfuse} used a LSTM to fuse bi-modal EHR data with modality missing case samples. In this paper, We reproduced this model as: 1) Bi-modal with Time-series and Images, and 2) Tri-modal with Time-series, Text, and Images.

    \item \textbf{Multi-modal Transformer (MT)}: \cite{ma2022multimodal} utilized the Transformer architecture with attention mask to fuse image and text with missing modality. 
    % We reproduced the model to handle tri-modal data: numeric, text, and images.

    \item \textbf{Multimodal Bottleneck Transformer(MBT)}: \cite{Nagrani2021AttentionBF} devised a modality-wise Transformer with bottleneck token to model inter-modality interaction. Since the original MBT can not receive data with modality missing, we develop our Skip Bottleneck (SB) to MBT for data with missing modality.
    % , which we will call Average Attention (AA) in this paper.
    % We reproduced the model to handle tri-modal data: numeric, text, and images.
    \end{itemize}

    \subsection{Robustness Against Missing Modality}
    We evaluate the model robustness against missing modality. To do so, we assess the model performance not only with the original test set but also with the test set with an increasing modality missing rate.
    In this study, we conduct three different experiments: 1) missing robustness of MBT with three different modality-attention scores, 2) training strategies to increase model robustness against modality missing, and 3) UMSE on missing robustness.
    First, we assess the missing robustness of TSA, CTAA, and AA as illustrated in Figure \ref{fig4}. Second, we explore two different strategies to increase missing robustness, i.e., Missing-Modal Augmentation (MMA) and layer optimization (Table \ref{tab4}) with multi-task learning (i.e., multi-token) as suggested in \cite{ma2022multimodal}. For layer optimization, we varied the fusion starting layer of TSA and selected the best fusion starting layer, which we call Fusion Layer Search (FLS), based on validation AUPRC. Moreover, we implemented the multi-token strategy as suggested by \cite{ma2022multimodal}. Lastly, we examined if UMSE contributes to the missing robustness based on TSA.

%% file: Sections/Table1_data.tex
% \begin{table}[h]
% \footnotesize
% \centering
% \caption{Data statistics with patient numbers for mortality prediction, vasopressor need and intubation need prediction tasks with modality missing information.}\label{dataset}
% \begin{tabular}{ccc}
% \hline
% \bfseries Tasks & \bfseries Mortality & \bfseries Vasopressor\\
% \hline
%   Data Split & Train / Test & Train / Test\\
% \hline
%   Positive Subjects & 2544 / 262 & 5827 / 606\\
%   Negative Subjects & 24492 / 2836 & 21941 / 2580\\
% \hline
%   \end{tabular}
%   \vspace{-15}
% \end{table}

% % *

 % X-ray image data is from MIMIC-CXR and chief-complaint text data is from MIMIC-ED. The data table with clinical note from MIMIC-IV-Note in the replacement of the chief-complaint is in Appendix\ref{tab6}
\begin{table}[!ht]
    \footnotesize
    \centering
    \renewcommand{\arraystretch}{0.92}
    \caption{Data statistics with the number of subjects for mortality prediction, vasopressor need, and intubation need prediction tasks with modality missing information.}\label{tab1}
    \begin{tabular}{l|ll|ll|ll}
    \toprule
        \multicolumn{7}{c}{\textbf{(a) Mortality Prediction}} \\  \hline
        {} & \multicolumn{2}{c}{Training} & \multicolumn{2}{c}{Validation} & \multicolumn{2}{c}{Test} \\ \cline{2-7}
        ~ & Positive & Negative & Positive & Negative & Positive & Negative \\ \hline
        Patient Number & 3486 & 30870 & 430 & 3741 & 413 & 3873 \\ \hline
        % \hspace*{3mm} Image: O, Text: O & 784 & 5338 & 106 & 597 & 101 & 652 \\ \hline
        % \hspace*{3mm} Image: X, Text: O & 1072 & 9909 & 125 & 1210 & 131 & 1269 \\ \hline
        % \hspace*{3mm} Image: O, Text: X & 80 & 1502 & 10 & 205 & 6 & 191 \\ \hline
        % \hspace*{3mm} Image: X, Text: X & 1550 & 14121 & 189 & 1729 & 175 & 1761 \\ \hline
        Image Missing Rate & 75.22\% & 77.84\% & 73.02\% & 78.56\% & 74.09\% & 78.23\% \\ \hline
        Text Missing Rate & 46.76\% & 50.61\% & 46.28\% & 51.70\% & 43.83\% & 50.40\% \\\bottomrule
    \end{tabular}

    \begin{tabular}{l|ll|ll|ll}
    % \toprule
        \multicolumn{7}{c}{\textbf{(b) Vasopressor Need Prediction}} \\  \hline
        {} & \multicolumn{2}{c}{Training} & \multicolumn{2}{c}{Validation} & \multicolumn{2}{c}{Test} \\ \cline{2-7}
        ~ & Positive & Negative & Positive & Negative & Positive & Negative \\ \hline
        Patient Number & 9341 & 24822 & 1172 & 2969 & 1183 & 3085  \\ \hline
        % \hspace*{3mm} Image: O, Text: O & 1966 & 4125 & 251 & 450 & 249 & 502  \\ \hline
        % \hspace*{3mm} Image: X, Text: O & 2767 & 8134 & 351 & 968 & 377 & 1014  \\ \hline
        % \hspace*{3mm} Image: O, Text: X & 528 & 1049 & 62 & 153 & 51 & 146  \\ \hline
        % \hspace*{3mm} Image: X, Text: X & 4080 & 11514 & 508 & 1398 & 506 & 1423  \\ \hline
        Image Missing Rate & 73.30\% & 79.16\% & 73.29\% & 79.69\% & 74.64\% & 79.00\%  \\ \hline
        Text Missing Rate & 49.33\% & 50.61\% & 48.63\% & 52.24\% & 47.08\% & 50.86\%  \\  \bottomrule
    \end{tabular}

    \begin{tabular}{l|ll|ll|ll}
    % \toprule
        \multicolumn{7}{c}{\textbf{(c) Intubation Need Prediction}} \\  \hline
        {} & \multicolumn{2}{c}{Training} & \multicolumn{2}{c}{Validation} & \multicolumn{2}{c}{Test} \\ \cline{2-7}
        ~ & Positive & Negative & Positive & Negative & Positive & Negative \\ \hline
        Patient Number & 13450 & 20682 & 1665 & 2467 & 1716 & 2552  \\ \hline
        % \hspace*{3mm} Image: O, Text: O & 2847 & 3237 & 321 & 376 & 349 & 401  \\ \hline
        % \hspace*{3mm} Image: X, Text: O & 2976 & 7912 & 368 & 950 & 415 & 978  \\ \hline
        % \hspace*{3mm} Image: O, Text: X & 823 & 752 & 120 & 95 & 103 & 94  \\ \hline
        % \hspace*{3mm} Image: X, Text: X & 6804 & 8781 & 856 & 1046 & 849 & 1079  \\ \hline
        Image Missing Rate & 72.71\% & 80.71\% & 73.51\% & 80.91\% & 73.66\% & 80.60\%  \\ \hline
        Text Missing Rate & 56.71\% & 46.09\% & 58.62\% & 46.25\% & 55.48\% & 45.96\%  \\ \bottomrule
    \end{tabular}
\end{table}

%% file: Sections/7_Results.tex
\section{Results}
    \label{sec:results}
    In this section, we report the predictive performance of 1) different multi-modal fusion models (Section \ref{sotaresults}) and 2) various strategies to enhance the robustness against missing modality (Section \ref{missingresult1}, \ref{missingresult2}) using the test AUPRC and AUROC, averaged over $3$ runs.
    
    % Our evaluation will proceed in the following steps: 1) comparing our proposed model to existing multi-modal models capable of managing missing modalities; 2) conducting a component analysis of SMSE; 3) exploring missing modality robustness by artificially increasing the absence of specific modalities.

    % 1. baseline 비교
    \input{Sections/Table2_result}
    \subsection{Comparison with State-of-the-art}
    \label{sotaresults}
    Table \ref{tab2} shows that our proposed models (i.e., TSA, CTAA, as described in Figure \ref{fig3}) score the highest predictive performance in all three clinical tasks. Specifically, TSA and CTAA exhibit the highest performance (AUPRC/AUROC) in tri-modal fusion; 0.864/0.970, 0.837/0.928, 0.868/0.934 in mortality, vasopressor need, and intubation need prediction. Moreover, TSA outperforms CTAA in mortality prediction, which may show the importance of EHR Time-series in mortality prediction. Note that EHR Image usually scores higher predictive performance than EHR Text with EHR Time-series, indicating that EHR Image provides more supplementary information for EHR Time-series in all predictive tasks.

    % 2. Set Embedding on different modalities (table 3)
    \input{Sections/Table3_SharedTime}
    \subsection{Set embedding benefits not only the EHR Time-series but also other EHR modalities}
    Table \ref{tab3} demonstrates that UMSE consistently enhances the performance in all three predictive tasks. Though UMSE increases the performance for all modalities, UMSE is more important for the EHR Time-series data. It is important to note that providing occurrence time for non-Time-series data too always enhances the performance in our clinical tasks, though it is usually disregarded in EHR studies.
    
    % We estimate the effectiveness of UMSE on missing modality inference using TSA by comparing performance when individual components are removed. The cases compared include: 1) no UMSE for time-series data, 2) no UMSE for image and text modality, and 3) no UMSE for all modalities.
    % , using conventional carry-forward imputation for EHR multimodal data preparation.

    % Performance with less modality in the test set
    %   - dataset 특화 설명
    %   - Robustness on missing modality with different attention strategies
    %   - Missing modality augmentation for robustness on missing EHR modality
    %   - UMSE for robustness on missing EHR modality
    \input{Sections/Figure4}
    \subsection{Missing EHR Image decreases the predictive performance more than missing EHR Text}
    \label{missingresult1}
    We compare the robustness against missing modality for three different modality-aware attention strategies (Figure \ref{fig4}). Note that the case percentage (x-axis) is the ratio from the whole test cases (i.e., 4286, 4268, 4268, see Table \ref{tab1}). As described in Table \ref{tab1}, the number of test cases with EHR Image is approximately 25\% in all three clinical tasks, and therefore, the maximum case percentage is approximately 25\%.
    When an equal number of cases lose their EHR Image and EHR Text, we can observe that losing EHR Image decreases the predictive performance more. Moreover, among three different clinical tasks, intubation need prediction may be more vulnerable to losing EHR Image indicated by the largest decaying slope for increasing EHR Image missing rate.

    \subsection{The optimal strategy for missing modality varies with the clinical tasks}
    \label{missingresult2}
    We compare two different strategies to increase the robustness against the modality missing (Figure \ref{fig5}): 1) Missing-Modality Augmentation (MMA) and 2) fusion layer optimization with multi-task learning as suggested in \cite{ma2022multimodal}. As shown in Figure \ref{fig5}, no single strategy excels. MMA outperforms others in missing EHR image for mortality and intubation need prediction, whereas \cite{ma2022multimodal} outperforms others in vasopressor need for missing either modality. It has to be noted that missing EHR text for mortality and intubation need prediction was not improved by either strategy.
    \input{Sections/Figure5}
    % models against the modality missing on the test set. We examined missing modality robustness from three perspectives:
    
    % \begin{itemize}
    % \item First, we compare text and image modality missing robustness with three different modality-aware attention strategies (Figure.\ref{fig4}).
    
    % \item Second, we explore missing modality robustness optimization methods, including Missing Modal Augmentation (MMA) and FSL with multi-task learning methods (FSL+Multi-Task) (\ref{fig5}).
    
    % \item Lastly, we assessed the efficacy of UMSE on missing modality robustness (\ref{fig6}).
    % \end{itemize}

    % umse를 통해 set encoding방식으로 데이터를 준비하는건 모든 모달리티에 중요함. 특히 time-series modality에게 umse를 제거 했을때, 성능하락이 심함. 이 경향성은 figure 6에서도 어떤 modality가 missing 되든지 consistent하게 나타남

%% file: Sections/Table2_result.tex
\begin{table*}[ht]
    \scriptsize
    \centering
    \caption{Performance comparison between baseline models and our proposed models for three clinical event prediction tasks, using bimodal (a, b) and trimodal (c) data.}\label{tab2}
    \begin{tabular}{l|ll|ll|ll}
    \hline
    %     \multicolumn{7}{c}{\textbf{EHR Times-Series}} \\  \hline
    %      & \multicolumn{2}{c}{Mortality} & \multicolumn{2}{c}{Vasopressor Need} & \multicolumn{2}{c}{Intubation Need} \\ \cline{2-7}
    %     ~ & AUPRC & AUROC & AUPRC & AUROC & AUPRC & AUROC \\ \hline
    %     LSTM &  &  & 0.698$\pm$0.004 & 0.883$\pm$0.001 & 0.580$\pm$0.002 & 0.805$\pm$0.0 \\ \hline
    %     GRUD &  &  & 0.720$\pm$0.007 & 0.893$\pm$0.003 & 0.618$\pm$0.005 & 0.813$\pm$0.002 \\ \hline
    %     Transformer & 0.704$\pm$0.0 & 0.928$\pm$0.0 & 0.694$\pm$0.02 & 0.885$\pm$0.0 & 0.597$\pm$0.0 & 0.815$\pm$0.01 \\ \hline
    %     ITE-Transformer & \textbf{0.809$\pm$0.0} & \textbf{0.967$\pm$0.0} & \textbf{0.741$\pm$0.0} & \textbf{0.903$\pm$0.0} & \textbf{0.761$\pm$0.0} & \textbf{0.909$\pm$0.0} \\ \hline
    % \midrule
        \multicolumn{7}{c}{\textbf{(a) EHR Times-Series + EHR Text}} \\  \hline
        & \multicolumn{2}{c}{Mortality} & \multicolumn{2}{c}{Vasopressor Need} & \multicolumn{2}{c}{Intubation Need} \\ \hline
        ~ & AUPRC & AUROC & AUPRC & AUROC & AUPRC & AUROC \\ \hline
        MNRIFN & 0.703$\pm$0.006 & 0.930$\pm$0.001 & 0.649$\pm$0.005 & 0.863$\pm$0.001 & 0.507$\pm$0.006 & 0.749$\pm$0.003 \\ \hline
        HAIM & 0.704$\pm$0.003 & 0.933$\pm$0.002 & 0.691$\pm$0.0 & 0.880$\pm$0.001 & 0.572$\pm$0.001 & 0.800$\pm$0.001 \\ \hline
        MT & 0.787$\pm$0.005 & 0.9564$\pm$0.001 & \textbf{0.751$\pm$0.001} & \textbf{0.906$\pm$0.001} & 0.739$\pm$0.012 & 0.902$\pm$0.005 \\ \hline
        ours (AA) & \textbf{0.801$\pm$0.004} & \textbf{0.961$\pm$0.002} & 0.743$\pm$0.002 & 0.901$\pm$0.002 & 0.744$\pm$0.009 & 0.903$\pm$0.005 \\ \hline
        ours (TSA) & 0.790$\pm$0.006 & 0.955$\pm$0.002 & 0.748$\pm$0.002 & 0.903$\pm$0.002 & \textbf{0.751$\pm$0.002} & \textbf{0.904$\pm$0.001} \\ \hline
        ours (CTAA) & 0.789$\pm$0.004 & 0.955$\pm$0.002 & 0.748$\pm$0.002 & 0.904$\pm$0.001 & 0.751$\pm$0.001 & 0.903$\pm$0.002 \\ \hline
    \midrule
        \multicolumn{7}{c}{\textbf{(b) EHR Times-Series + EHR Image}} \\  \hline
        & \multicolumn{2}{c}{Mortality} & \multicolumn{2}{c}{Vasopressor Need} & \multicolumn{2}{c}{Intubation Need} \\ \hline
        ~ & AUPRC & AUROC & AUPRC & AUROC & AUPRC & AUROC \\ \hline
        Medfuse  & 0.626$\pm$0.005 & 0.924$\pm$0.003 & 0.698$\pm$0.004 & 0.884$\pm$0.001 & 0.583$\pm$0.002 & 0.802$\pm$0.001 \\ \hline
        HAIM & 0.696$\pm$0.007 & 0.927$\pm$0.004 & 0.703$\pm$0.001 & 0.885$\pm$0.0 & 0.579$\pm$0.002 & 0.802$\pm$0.001 \\ \hline
        MT & 0.799$\pm$0.005 & 0.96$\pm$0.002 & 0.778$\pm$0.005 & 0.915$\pm$0.004 & 0.794$\pm$0.006 & 0.908$\pm$0.0 \\ \hline
        ours (AA) & 0.807$\pm$0.006 & 0.963$\pm$0.003 & 0.78$\pm$0.007 & 0.915$\pm$0.0 & 0.803$\pm$0.004 & 0.913$\pm$0.002 \\ \hline
        ours (TSA) & \textbf{0.811$\pm$0.01} & \textbf{0.963$\pm$0.003} & 0.777$\pm$0.003 & 0.916$\pm$0.002 & \textbf{0.807$\pm$0.001} & \textbf{0.915$\pm$0.001} \\ \hline
        ours (CTAA) & 0.805$\pm$0.001 & 0.962$\pm$0.002 & \textbf{0.781$\pm$0.002} & \textbf{0.916$\pm$0.001} & 0.804$\pm$0.006 & 0.915$\pm$0.0 \\ \hline
    \midrule
        \multicolumn{7}{c}{\textbf{(c) EHR Times-Series + EHR Text + EHR Image}} \\  \hline
        & \multicolumn{2}{c}{Mortality} & \multicolumn{2}{c}{Vasopressor Need} & \multicolumn{2}{c}{Intubation Need} \\ \hline
        ~ & AUPRC & AUROC & AUPRC & AUROC & AUPRC & AUROC \\ \hline
        Medfuse & 0.819$\pm$0.007 & 0.955$\pm$0.002 & 0.800$\pm$0.001 & 0.912$\pm$0.001 & 0.775$\pm$0.001 & 0.871$\pm$0.0 \\ \hline
        HAIM & 0.699$\pm$0.007 & 0.931$\pm$0.003 & 0.687$\pm$0.004 & 0.881$\pm$0.002 & 0.575$\pm$0.005 & 0.798$\pm$0.001 \\ \hline
        MT & 0.844$\pm$0.004 & 0.961$\pm$0.005 & 0.831$\pm$0.003 & 0.926$\pm$0.001 & 0.854$\pm$0.003 & 0.927$\pm$0.001 \\ \hline
        ours (AA) & 0.847$\pm$0.002 & 0.965$\pm$0.003 & 0.831$\pm$0.006 & 0.926$\pm$0.004 & 0.861$\pm$0.003 & 0.931$\pm$0.0 \\ \hline
        ours (TSA) & \textbf{0.864$\pm$0.01} & \textbf{0.970$\pm$0.005} & 0.836$\pm$0.001 & 0.929$\pm$0.001 & 0.860$\pm$0.003 & 0.931$\pm$0.0 \\ \hline
        ours (CTAA) & 0.854$\pm$0.005 & 0.964$\pm$0.001 & \textbf{0.837$\pm$0.004} & \textbf{0.928$\pm$0.004} & \textbf{0.868$\pm$0.003} & \textbf{0.934$\pm$0.002} \\ \hline
    \end{tabular}
\end{table*}

 % $TS$ refers to Times-Series modality. $Img$ refers to X-ray image modality. $Txt$ refers to EHR text modality.

%% file: Sections/Table3_SharedTime.tex
\begin{table*}[ht]
    \scriptsize
    \centering
    \caption{Performance analysis of TSA based on the application of UMSE on time-series and non-time-series modalities.}\label{tab3}
    \begin{tabular}{l|ll|ll|ll}
    \hline
        % \multicolumn{7}{c}{\textbf{Unimodal Models (EHR Times-Series)}} \\  \hline
         & \multicolumn{2}{c}{\textbf{Mortality}} & \multicolumn{2}{c}{\textbf{Vasopressor Need}} & \multicolumn{2}{c}{\textbf{Intubation Need}} \\ \cline{2-7}
        ~ & AUPRC & AUROC & AUPRC & AUROC & AUPRC & AUROC \\ \hline
        Remove Time-Series UMSE & 0.80$\pm$0.01 & 0.94$\pm$0.0 & 0.79$\pm$0.0 & 0.91$\pm$0.0 & 0.77$\pm$0.0 & 0.87$\pm$0.0 \\ \hline
        Remove Text Image UMSE & 0.84$\pm$0.0 & 0.96$\pm$0.0 & 0.83$\pm$0.0 & 0.92$\pm$0.0 & 0.85$\pm$0.0 & 0.93$\pm$0.0 \\ \hline
        Remove all UMSE & 0.79$\pm$0.01 & 0.93$\pm$0.01 & 0.78$\pm$0.0 & 0.90$\pm$0.0 & 0.77$\pm$0.01 & 0.86$\pm$0.0 \\ \hline
        \midrule
        Default (TSA) & \textbf{0.86$\pm$0.01} & \textbf{0.97$\pm$0.01} & \textbf{0.84$\pm$0.0} & \textbf{0.93$\pm$0.0} & \textbf{0.86$\pm$0.0} & \textbf{0.93$\pm$0.0} \\ \hline

    \end{tabular}
\end{table*}

%% file: Sections/Figure4.tex
\begin{figure}[t]
\centering
% \vspace{-12}
\includegraphics[width=\textwidth]{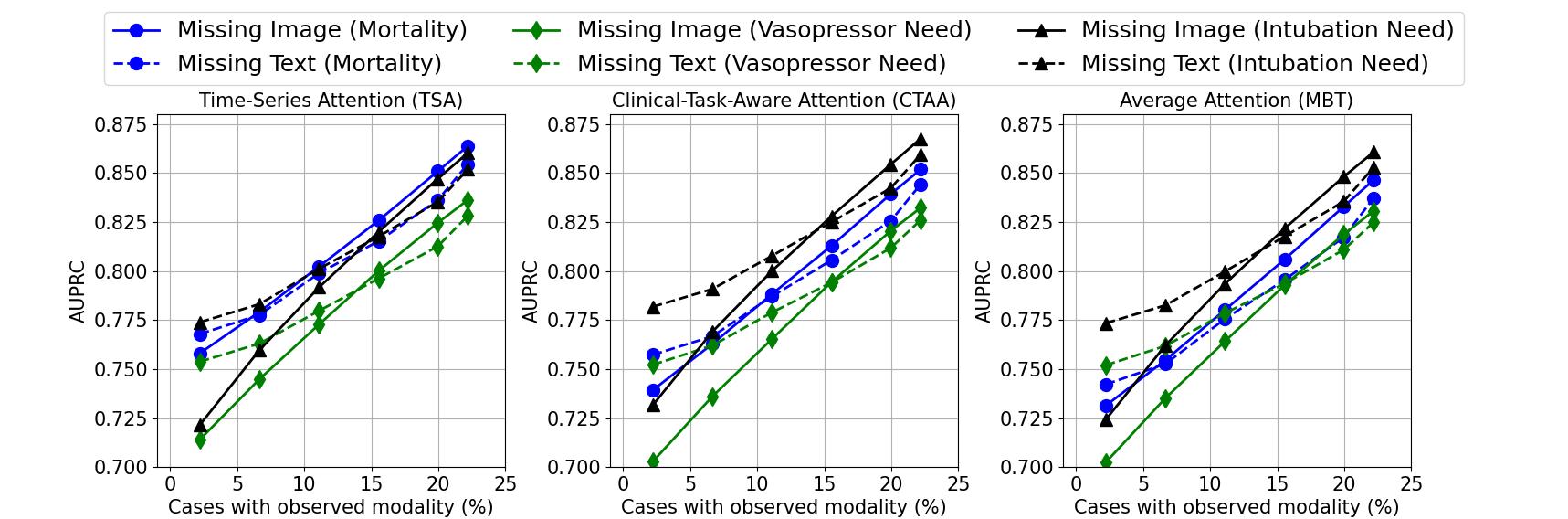}
\caption{Missing modal robustness with different modality-aware attention methods.}
\label{fig4}
% \vspace{-12}
\end{figure}

%% file: Sections/Figure5.tex
\begin{figure}[hb!]
% \centering
% \vspace{-12}
\includegraphics[width=\textwidth]{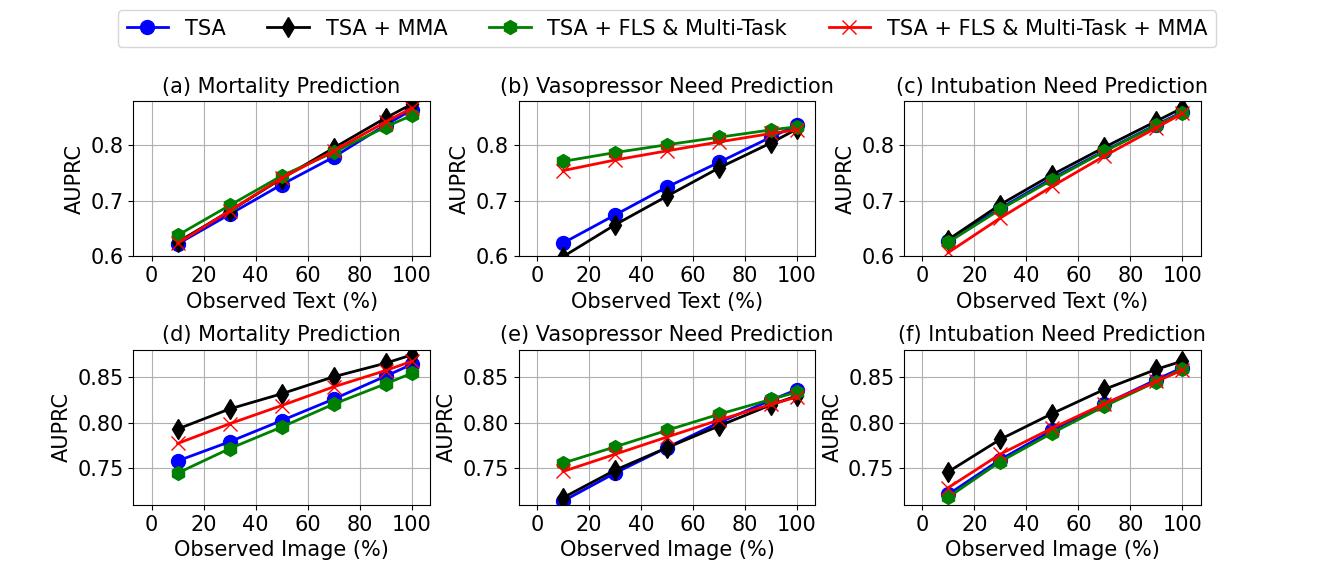}
\caption{Result of the robustness against missing modalities for various training strategies, i.e., MMA, FLS with Multi-Task, and Both.}
\label{fig5}
% \vspace{-12}
\end{figure}

%% file: Sections/8_Discussion.tex
\input{Sections/Figure6}
\section{Discussion}
    \label{sec:discussion}
    
    % UMSE missing (fig7). UMSE는 모든 모달리티에 필요하다
    % softmax (fig6) -> mortality는 time-series 가 중요함 <- supported by the outperformance of TSA
    % layer fusion
    % tri-modal 제일 좋아, consistently with all algorithms (tab2)
    % Transformer-based model는 consistent 한 EHR multi-modal fusion strategy 다.
    % UMSE는 이미지를 여러개 가져올 수 있지만, 여러개 가져와 봤자 도움이 안되고 아예 없으면 안좋다. text는 1440 시간동안 환자당 하나밖에 없어서 두개 이상 가져올수없었다,

    \subsection{Both Skip Bottleneck and Modality-Aware Attention modules improve learning multi-modal EHR}
    All of our models (i.e., TSA, CTAA, AA) are built on top of Skip Bottleneck since MBT does not consider modality missing. According to Table \ref{tab2}, all our models outperform other algorithms (including MT) except in vasopressor need prediction with bi-modal EHR. Our additional modality-aware attention scheme (TSA, CTAA) further improves the predictive performance, especially in mortality prediction. Note that the mortality prediction depends more heavily on EHR Time-series than other modalities and clinical tasks as illustrated in Figure \ref{fig7}. We assume that the Skip Bottleneck with the default attention from MBT (AA) was outperformed by models with additional modality-aware attention scheme especially in mortality prediction because mortality prediction is the clinical task with the most biased modality dependency as illustrated in Figure \ref{fig7}. Note that according to Figure \ref{fig7}, vasopressor need prediction task does not depend on a single modality whereas the intubation need prediction task makes slightly more attention to EHR Text than to other modalities.
    
    % 강조점 1: Transformer기반 fusion model들이 전부 각 테스크에서 1등을 함. 이유는 mnrifn, HAIM, medfuse처럼 단순 late fusion보다 mid fusion으로 fusion이 더 잘됐을것으로 추측.
    \input{Sections/Table4_layerfusion.tex}
    \input{Sections/Figure7}
    % \subsection{Late fusion strategy can be detrimental to learning multi-modal EHR}
    % As illustrated in Table \ref{tab2}, the Transformer-based fusion algorithms, which are all early fusion by default, excel other alternatives in all three clinical tasks (all other alternatives adopt a late fusion strategy). Moreover, in Table \ref{tab4}, the late fusion strategy drastically decreases the predictive performance in all clinical tasks though the base algorithm was a Transformer-based algorithm (TSA). From these two observations, we can conjecture that late fusion is detrimental to fusing multi-modal EHR data.
    \subsection{Late fusion strategy can be detrimental to learning multi-modal EHR}
    As illustrated in Table \ref{tab2}, the Transformer-based fusion algorithms, which are all early fusion by default, excel other alternatives in all three clinical tasks (all other alternatives adopt a late fusion strategy). Moreover, in Table \ref{tab4}, the late fusion strategy drastically decreases the predictive performance in all clinical tasks. From these two observations, we can conjecture that late fusion is detrimental to fusing multi-modal EHR data.
    
    % In Table \ref{tab2}, among bimodal models, those incorporating EHR image data show higher performance than those with EHR text data, suggesting that EHR images provide more valuable information for clinical event prediction than EHR chief-complaint text data.

    \subsection{Unified Multi-modal Set Embedding benefits multi-modal EHR learning}
    In addition to Table \ref{tab3}, Figure \ref{fig6} also demonstrates that UMSE consistently enhances EHR multi-modal learning in all clinical tasks. This finding highlights that Unified Multi-modal Set Embedding (UMSE) method is advantageous not only for learning EHR time-series data but also for modeling other EHR modalities such as EHR Images and EHR Text. The performance gain is consistent across all three clinical tasks and all missing modality rates.

    % \subsection{Image modality benefits clinical intervention need prediction more than text modality}
    % In clinical intervention prediction, transformer models gain greater advantages from image modality than text modality. Despite the higher missing rate in image data (Table \ref{tab1}), bimodal fusion models with image modality surpass those with text modality, achieving a 3\% AUPRC increase in vasopressor need prediction and a 5.6\% AUPRC increase in intubation need prediction (Table \ref{tab2}). This outcome implies that image modality data offers more valuable information for predicting clinical interventions compared to text modality.

    \subsection{Limitations}
    Our main focus is on 1) a unified set embedding for all EHR modalities and 2) modality-aware attention. We assess the robustness of our algorithms against missing modality (Figure \ref{fig4}, Figure \ref{fig5}, and Figure \ref{fig6}) with recently introduced approaches to tackle missing modality \cite{ma2022multimodal}. However, we hardly found a single solution to improve the modality-missing problem. We also discovered that no approach could improve the robustness on missing EHR Text in mortality and intubation need prediction. As a result, we believe that the research on a missing modality will be fitting to our following research.

%% file: Sections/Figure6.tex
\begin{figure}[ht]
% \centering
% \vspace{-12}
\includegraphics[width=\textwidth]{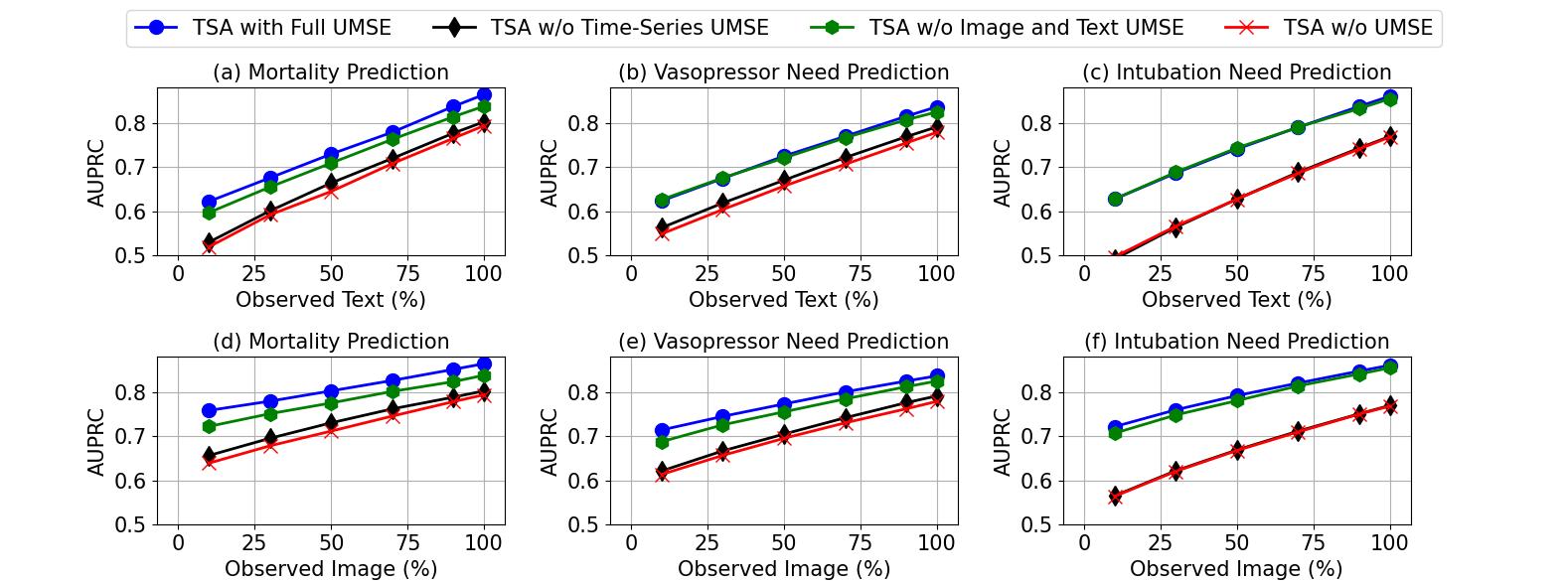}
\caption{Result of the robustness against missing modalities for modality-wise UMSE implementation.}
\label{fig6}
% \vspace{-12}
\end{figure}

%% file: Sections/Table4_layerfusion.tex
\begin{table*}[ht]
    \scriptsize
    \centering
    \caption{Result of optimal fusion layer search using TSA. The performance is measured in AUPRC and AUROC of validation dataset.}\label{tab4}
    \begin{tabular}{lllllll}
    \hline
        % \multicolumn{7}{c}{\textbf{Unimodal Models (EHR Times-Series)}} \\  \hline
         & \multicolumn{2}{c}{\textbf{Mortality}} & \multicolumn{2}{c}{\textbf{Vasopressor Need}} & \multicolumn{2}{c}{\textbf{Intubation Need}} \\\hline
        $L_{fusion}$ & AUPRC & AUROC & AUPRC & AUROC & AUPRC & AUROC \\ \hline
        1 & 0.861$\pm$0.007 & 0.968$\pm$0.003 & 0.843$\pm$0.003 & 0.929$\pm$0.001 & 0.862$\pm$0.001 & 0.932$\pm$0.001 \\ \hline
        2 & 0.858$\pm$0.006 & 0.967$\pm$0.002 & 0.842$\pm$0.003 & 0.928$\pm$0.003 & 0.865$\pm$0.003 & 0.933$\pm$0.0 \\ \hline
        3 & \textbf{0.862$\pm$0.005} & \textbf{0.966$\pm$0.004} & 0.841$\pm$0.001 & 0.928$\pm$0.002 & \textbf{0.866$\pm$0.004} & \textbf{0.933$\pm$0.001} \\ \hline
        4 & 0.858$\pm$0.003 &  0.968$\pm$0.002 & 0.840$\pm$0.002 & 0.927$\pm$0.003 & 0.864$\pm$0.005 & 0.932$\pm$0.003 \\ \hline
        5 & 0.857$\pm$0.007 & 0.969$\pm$0.0 & \textbf{0.843$\pm$0.001} & \textbf{0.928$\pm$0.0} & 0.865$\pm$0.001 & 0.933$\pm$0.002 \\ \hline
        6 & 0.818$\pm$0.007 & 0.965$\pm$0.002 & 0.766$\pm$0.002 & 0.913$\pm$0.001 & 0.759$\pm$0.002 & 0.907$\pm$0.002 \\ \hline

    \end{tabular}
\end{table*}

%% file: Sections/Figure7.tex
\begin{figure}[ht]
\centering
% \vspace{-12}
\includegraphics[width=\textwidth]{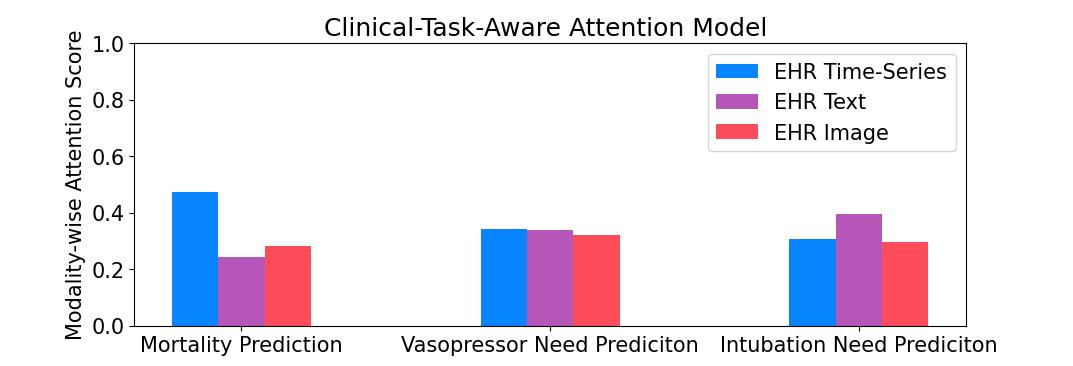}
\caption{Modality-wise attention score from our proposed CTAA model.}
\label{fig7}
% \vspace{-12}
\end{figure}

%% file: Sections/9_Appendix.tex
\appendix
\section{Detailed information on Experimental Setting}
%
% 전처리 및 라벨링 설명
%
\subsection{Cohort}
A total of 53,150 patients with valid ICU admission and age over 18 were identified in the MIMIC-IV dataset. After excluding 20 patients due to missing vital sign records (i.e., pulse, systolic blood pressure, diastolic blood pressure, respiratory rate, and body temperature), the final cohort consists of 53,130 subjects. We further excluded patients using the process described in Section \ref{sec:datapreprocess}. Finally, for clinical event and intervention need prediction tasks, we followed the process illustrated in Section \ref{sec:taskdataset}. In the end, we collect 42,813, 42,572, and 42,532 ICU subjects for mortality, vasopressor need, and intubation need prediction tasks respectively.

\subsubsection{Clinical Objectives}
\label{appendixtasks}
In MIMIC-IV, we categorize Intubation (224385(which is item ID)), Intubation - Details (223059), Oral ETT (225307), Nasal ETT (225308), Unplanned Extubation (patient-initiated) (225468), Unplanned Extubation (non-patient initiated) (225477), Timeout Performed by (Intubation) (226188) as intubation task and Norepinephrine (221906), Dopamine (221662), Dobutamine (221289), Epinephrine (221289) as vasopressor task.

\subsection{Data Sampler}
\label{datasampler}
As we implement real-time training incorporating up to 1440 hours per patient EHR data, there are an excessive amount of negatives. To address this imbalance during training, we employ a data sampler to impose equal proportions of positives and negatives during training.
%
% 전처리 for time-series data
%
\subsection{EHR Time-Series Data Preparation}
\label{ehrtimedata}
We select six vital-sign features, i.e., heart rate, respiration rate, diastolic and systolic blood pressure, temperature, and pulse oximetry. We gather ten features of laboratory result data, i.e., Hematocrit, Platelet, WBC, Bilirubin, pH, HCO3, Creatinine, Lactate, Potassium, and Sodium (\cite{sung2021event}). 

%
% 전처리 for text data
%
\subsection{EHR Text Data Preparation}
\label{ehrtextdata}
\subsubsection{Chief-complaint from MIMIC-ED}
We use the chief complaint from MIMIC-ED as our EHR text data. When a chief complaint is available, it indicates that the subject has visited the Emergency Department (ED) prior to ICU admission.

\subsubsection{Clinical Notes from MIMIC-IV-Note}

\input{Sections/Table5_clinic_datat}
%1. Chief Complaint, 2. Past Medical History, 3. Medications on admission (초기 정보) 
We extract three sections of clinical notes: Chief Complaint, Past Medical History, and Medications on admission. We focus on these sections since we only require initial subject information to predict the occurrence of specific tasks within the hospital. To separate the extracted sections of clinical notes, we used [SEP] tokens. If the notes do not contain any of the three sections and if there were no available substitutes (such as Medical/Surgical History instead of Past Medical History), we only used the [SEP] token as the section separator. We exclude optional information about Past Medical History, such as Other Past Medical History, and also removed Medications-OTC from the Medications on Admission category, as it is not useful for predicting our tasks. The data table for the clinical note replacing chief complaint as our EHR text data is illustrated in Table \ref{tab5}, and the experimental results are illustrated in \ref{clinicalnoteresult}. Note that the missing rate of the chief complaint is smaller than the clinical note (Table \ref{tab1}).

\subsubsection{Pre-trained Text Embedder}
\label{pretrainedtext}
We employed BioBERT, a pre-trained biomedical language representation model, to encode chief complaints or clinical notes from MIMIC-ED or MIMIC-IV-Note. The maximum lengths were set to 128 and 512 for chief complaints and clinical notes, respectively. During our predictive training, BioBERT provided embeddings for both text data as a sequence of tokens. The pre-trained BioBERT utilized a vocabulary size of 28,996 to generate the sequence of embedding vectors.

%
% 전처리 for image data
%
\subsection{EHR Image Data Preparation}
\label{ehrimagedata}
\subsubsection{Data Augmentations}
% 문장 느낌 MEdfuse 고쳐야 함, 차이점: horizontal flip 안함, random affine degree 45 -> 5, 논문에서는 train시 random crop but github에서는 center crop, 우리는 center crop
To augment chest X-ray images, we applied a series of transformations during both the pre-training and fine-tuning phases. Specifically, we 1) resized each image to 256 × 256 pixels, 2) employed a set of random affine transformations, i.e., rotation, scaling, and translation, and 3) performed center-crop to obtain an image of size 224 × 224 pixels. For validation and testing, the images were resized to 256 × 256 pixels and underwent the same center crop operation to obtain an image of size 224 × 224 pixels. We consistently applied these procedures throughout our experiments.
\subsubsection{Pre-trained Image Embedder}
\label{pretrainedimage}
We selected our pre-trained SwinTransformer as an EHR image embedder (\cite{liu2021swin}). We pre-trained Swin Transformer using CheXpert with the initial pre-trained weight from ImageNet-1K; we classify 14 binary radiology labels that were extracted from radiology reports via CheXpert(\cite{DBLP:journals/corr/abs-1901-07031}). We optimized SwinTransformer using the binary cross entropy loss with learning rate sweep from $10^{-6}$ to $10^{-4}$.
% It can hence be independently pre-trained using their respective labels and losses. 
% and selected the model with highest average AUROC. 
% and learning rate that achieve the best AUROC on the respective validation set.($10^{-6}$: 0.7527, $10^{-5}$: 0.7519, $10^{-4}$: 0.7493) We fix learning rate scheduler with Cosine Annealing Warm up Restarts.
To create pre-training dataset, we extracted chest X-ray images of ICU patients from MIMIC-CXR and randomly split them by patient ID. Our training, validation, and test set consists of 213,016, 23,131, and 26,744 images, respectively.
% The Swin Transformer (\cite{liu2021swin}) constructs hierarchical feature maps, offering flexibility to model at different scales while maintaining linear computational complexity relative to image size. 
We employed the tiny version of the SwinTransformer, which uses 96 feature dimension, a patch size of $4 \times 4$, a window size of $7 \times 7$, and block depths of [2, 2, 6, 2].

% Swin Transformer(\cite{liu2021swin}) builds hierarchical feature maps to have the flexibility to model at various scales and have linear computational complexity with respect to image size. We used tiny version of Swin Transformer. We utilized a model with embedding dimension of 96, patch size of $4 \times 4$, and window size of $7 \times 7 $, and depths of blocks used are [2,2,6,2]. 
% banner swin_chex_appa 
%Swin_T_Weights.IMAGENET1K_V1
% BCEwithlogit loss 사용, output 14, (layer, node 등)
% dim_output, node, layer
%문장 표절 안 뜨게 조정 필요
% 안필요 할 것 같지만... 쓸모 없는 특수문자 제거 (단, %, / , & 와 같은 단위나 약자를 나타낼 수 있는 것은 삭제 하지 않음)

\section{Supplementary Results}
\subsection{Real-Time Time Embedding}
    \label{realtimeresults}
    In UMSE, we employ the time difference between the occurrence time and the current time, calculated as $t_{occurrence}-t_{current}$. Table \ref{tab6} compares the predictive performance between using $t_{occurrence}-t_{current}$ and using $t_{occurrence}$ alone. Note that we calculated $t_{occurrence}$ as the time from the admission.
\input{Sections/Table6_realtime}

\subsection{Pre-training}
    \label{pretrainingresult}
    We employed pretrained Swin Transformer fine-tuned on the Chexpert dataset and BioBERT pretrained with PubMed 1M \footnote{\url{https://huggingface.co/dmis-lab/biobert-base-cased-v1.2}}. In this section, we present the predictive performances when each pretrained encoder is substituted with an ImageNet pretrained Swin Transformer and BERT Tokenizer (see Table \ref{tab7}).
\input{Sections/Table7_pretraining}

\subsection{Multiple X-rays (EHR Image) per patient}
    \label{multiimages}
    UMSE can differentiate multiple non-time series data, therefore enabling to use multiple images/texts per patient. Since a single subject only possesses single text, we evaluated if using multiple X-ray images is beneficial. Specifically, we experimented TSA model training with up to the most recent three X-ray images whose results are summarized in Table \ref{tab8}. Note that for intubation need prediction task, the prediction performance slightly increases when using multiple images.
\input{Sections/Table8_multimages.tex}

\subsection{Clinical Note for EHR text}
    \label{clinicalnoteresult}
    A recent release of the clinical note for the MIMIC-IV dataset enables us to compare the benefits of chief complaint versus clinical note, as detailed in Appendix \ref{ehrtextdata}. Note the slight performance increase by using clinical notes instead of the chief complaint for both mortality and vasopressor use prediction (Table \ref{tab9}).
\input{Sections/Table9_ClinicalNote}

%% file: Sections/Table5_clinic_datat.tex
\begin{table}[!ht]
    \footnotesize
    \centering
    \caption{Data statistics with patient numbers for mortality prediction, vasopressor need and intubation need prediction tasks with modality missing information. X-ray image is from MIMIC-CXR and clinical note text is from MIMIC-IV-Note.}\label{tab5}%tab6
    \begin{tabular}{l|ll|ll|ll}
    \toprule
        \multicolumn{7}{c}{\textbf{(a) Mortality Prediction}} \\  \hline
        {} & \multicolumn{2}{c}{Training} & \multicolumn{2}{c}{Validation} & \multicolumn{2}{c}{Test} \\ \hline
        ~ & Positive & Negative & Positive & Negative & Positive & Negative \\ \hline
        Patient Number & 3486 & 30870 & 430 & 3741 & 413 & 3873 \\ \hline
        Image Missing Rate & 75.22\% & 77.84\% & 73.02\% & 78.56\% & 74.09\% & 78.23\% \\ \hline
        Text Missing Rate & 10.01\% & 5.58\% & 6.28\% & 5.91\% & 8.23\% & 5.63\% \\\bottomrule
    \end{tabular}

    \begin{tabular}{l|ll|ll|ll}
    % \toprule
        \multicolumn{7}{c}{\textbf{(b) Vasopressor Need Prediction}} \\  \hline
        {} & \multicolumn{2}{c}{Training} & \multicolumn{2}{c}{Validation} & \multicolumn{2}{c}{Test} \\ \hline
        ~ & Positive & Negative & Positive & Negative & Positive & Negative \\ \hline
        Patient Number & 9341 & 24822 & 1172 & 2969 & 1183 & 3085  \\ \hline
        Image Missing Rate & 73.30\% & 79.16\% & 73.29\% & 79.69\% & 74.64\% & 79.00\%  \\ \hline
        Text Missing Rate & 6.22\% & 5.78\% & 7.0\% & 5.56\% & 5.75\% & 5.8\%  \\  \bottomrule
    \end{tabular}

    \begin{tabular}{l|ll|ll|ll}
    % \toprule
        \multicolumn{7}{c}{\textbf{(c) Intubation Need Prediction}} \\  \hline
        {} & \multicolumn{2}{c}{Training} & \multicolumn{2}{c}{Validation} & \multicolumn{2}{c}{Test} \\ \hline
        ~ & Positive & Negative & Positive & Negative & Positive & Negative \\ \hline
        Patient Number & 13450 & 20682 & 1665 & 2467 & 1716 & 2552  \\ \hline
        Image Missing Rate & 72.71\% & 80.71\% & 73.51\% & 80.91\% & 73.66\% & 80.60\%  \\ \hline
        Text Missing Rate & 5.87\% & 5.94\% & 6.43\% & 5.63\% & 6.43\% & 5.63\%  \\ \bottomrule
    \end{tabular}
\end{table}

%% file: Sections/Table6_realtime.tex
\begin{table*}[ht]
    \scriptsize
    % \centering
    \caption{\textbf{Effectiveness of Real-Time embedding} for three clinical event and intervention predictive tasks. The performance is measured in AUPRC and AUROC of validation dataset, averaged over $3$ runs.}\label{tab6}%tab9
    \begin{tabular}{l|ll|ll|ll}
    \hline
        % \multicolumn{7}{c}{\textbf{Unimodal Models (EHR Times-Series)}} \\  \hline
         & \multicolumn{2}{c}{Mortality} & \multicolumn{2}{c}{Vasopressor Need} & \multicolumn{2}{c}{Intubation Need} \\ \cline{2-7}
        ~ & AUPRC & AUROC & AUPRC & AUROC & AUPRC & AUROC \\ \hline
        No Real-Time & 0.828$\pm$0.005 & 0.961$\pm$0.003 & 0.834$\pm$0.001 & 0.926$\pm$0.001 & 0.855$\pm$0.003 & 0.928$\pm$0.001 \\ \hline
        Default (TSA) & \textbf{0.861$\pm$0.007} & \textbf{0.968$\pm$0.003} & \textbf{0.843$\pm$0.003} & \textbf{0.929$\pm$0.001} & \textbf{0.862$\pm$0.001} & \textbf{0.932$\pm$0.001} \\ \hline

    \end{tabular}
\end{table*}

%% file: Sections/Table7_pretraining.tex
\begin{table*}[ht]
    \scriptsize
    % \centering
    \caption{\textbf{Result of Pretrained Encoder Effectiveness} for three clinical event and intervention predictive tasks. Here, INP refers to ImageNet pretrained Swin Transformer instead of Chexpert label pretrained Swin Transformer. BT refers to BERT tokenzier instead of BioBERT used in TSA.}\label{tab7}%tab8
    \begin{tabular}{l|ll|ll|ll}
    \hline
        % \multicolumn{7}{c}{\textbf{Unimodal Models (EHR Times-Series)}} \\  \hline
         & \multicolumn{2}{c}{Mortality} & \multicolumn{2}{c}{Vasopressor Need} & \multicolumn{2}{c}{Intubation Need} \\ \cline{2-7}
        ~ & AUPRC & AUROC & AUPRC & AUROC & AUPRC & AUROC \\ \hline
        TSA with INP & 0.849$\pm$0.004 &  0.965$\pm$0.002 & 0.843$\pm$0.004 & 0.927$\pm$0.003 & 0.855$\pm$0.003 & 0.924$\pm$0.002 \\ \hline
        TSA with BT & 0.845$\pm$0.003 & 0.961$\pm$0.002 & 0.839$\pm$0.003 & 0.926$\pm$0.003 & \textbf{0.863$\pm$0.002} & \textbf{0.931$\pm$0.003} \\ \hline
        \midrule
        Default (TSA) & \textbf{0.861$\pm$0.007} & \textbf{0.968$\pm$0.003} & \textbf{0.843$\pm$0.003} & \textbf{0.929$\pm$0.001} & 0.862$\pm$0.001 & 0.932$\pm$0.001 \\ \hline
        
    \end{tabular}
\end{table*}

%% file: Sections/Table8_multimages.tex
\begin{table*}[ht]
    \scriptsize
    \centering
    \caption{Result of TSA model when maximum three of most recent X-ray images are used. The performance is measured in AUPRC and AUROC of validation dataset, averaged over $3$ runs.}\label{tab8} %table7
    \begin{tabular}{lllllll}
    \hline
        % \multicolumn{7}{c}{\textbf{Unimodal Models (EHR Times-Series)}} \\  \hline
         & \multicolumn{2}{c}{Mortality} & \multicolumn{2}{c}{Vasopressor Need} & \multicolumn{2}{c}{Intubation Need} \\\hline
        $L_{fusion}$ & AUPRC & AUROC & AUPRC & AUROC & AUPRC & AUROC \\ \hline
        TSA (Default) & \textbf{0.86$\pm$0.01} & \textbf{0.96$\pm$0.0} & \textbf{0.84$\pm$0.0} & \textbf{0.93$\pm$0.0} & 0.86$\pm$0.0 & 0.93$\pm$0.0 \\ \hline
        TSA (maximum 3 images) & 0.85$\pm$0.0 & 0.96$\pm$0.0 & \textbf{0.84$\pm$0.0} & \textbf{0.93$\pm$0.0} & \textbf{0.87$\pm$0.0} & \textbf{0.94$\pm$0.0} \\ \hline
        
    \end{tabular}
\end{table*}

%% file: Sections/Table9_ClinicalNote.tex
\begin{table*}[ht]
    \scriptsize
    \centering
    \caption{Result of TSA model when MIMIC-IV-Note text data is utilized in the replacement of chief-complaint text from MIMIC-ED. The performance is measured in AUPRC and AUROC of validation dataset, averaged over $3$ runs.}\label{tab9}%table5
    \begin{tabular}{lllllll}
    \hline
        % \multicolumn{7}{c}{\textbf{Unimodal Models (EHR Times-Series)}} \\  \hline
         & \multicolumn{2}{c}{Mortality} & \multicolumn{2}{c}{Vasopressor Need} & \multicolumn{2}{c}{Intubation Need} \\\hline
        $L_{fusion}$ & AUPRC & AUROC & AUPRC & AUROC & AUPRC & AUROC \\ \hline
        With chief-complaint & 0.861$\pm$0.007 & 0.968$\pm$0.003 & 0.843$\pm$0.003 & 0.929$\pm$0.001 & \textbf{0.862$\pm$0.001} & \textbf{0.932$\pm$0.001} \\ \hline
        With clinical note & \textbf{0.864$\pm$0.007} & \textbf{0.971$\pm$0.002} & \textbf{0.851$\pm$0.001} & \textbf{0.934$\pm$0.0} & 0.855$\pm$0.002 & 0.928$\pm$0.001 \\ \hline
        
    \end{tabular}
\end{table*}

%% file: mlhc.bbl
\begin{thebibliography}{36}
\providecommand{\natexlab}[1]{#1}
\providecommand{\url}[1]{\texttt{#1}}
\expandafter\ifx\csname urlstyle\endcsname\relax
  \providecommand{\doi}[1]{doi: #1}\else
  \providecommand{\doi}{doi: \begingroup \urlstyle{rm}\Url}\fi

\bibitem[Akbari et~al.(2021)Akbari, Yuan, Qian, Chuang, Chang, Cui, and
  Gong]{akbari2021vatt}
Hassan Akbari, Liangzhe Yuan, Rui Qian, Wei-Hong Chuang, Shih-Fu Chang, Yin
  Cui, and Boqing Gong.
\newblock Vatt: Transformers for multimodal self-supervised learning from raw
  video, audio and text.
\newblock \emph{Advances in Neural Information Processing Systems},
  34:\penalty0 24206--24221, 2021.

\bibitem[Alayrac et~al.(2020)Alayrac, Recasens, Schneider, Arandjelovi'c,
  Ramapuram, Fauw, Smaira, Dieleman, and
  Zisserman]{Alayrac2020SelfSupervisedMV}
Jean-Baptiste Alayrac, Adri{\`a} Recasens, Rosalia Schneider, Relja
  Arandjelovi'c, Jason Ramapuram, Jeffrey~De Fauw, Lucas Smaira, Sander
  Dieleman, and Andrew Zisserman.
\newblock Self-supervised multimodal versatile networks.
\newblock \emph{ArXiv}, abs/2006.16228, 2020.

\bibitem[Che et~al.(2018)Che, Purushotham, Cho, Sontag, and
  Liu]{che2018recurrent}
Zhengping Che, Sanjay Purushotham, Kyunghyun Cho, David Sontag, and Yan Liu.
\newblock Recurrent neural networks for multivariate time series with missing
  values.
\newblock \emph{Scientific reports}, 8\penalty0 (1):\penalty0 6085, 2018.

\bibitem[Choi et~al.(2022)Choi, Chung, Chung, Lee, Hyun, and
  Kim]{choi2022advantage}
Arom Choi, Kyungsoo Chung, Sung~Phil Chung, Kwanhyung Lee, Heejung Hyun, and
  Ji~Hoon Kim.
\newblock Advantage of vital sign monitoring using a wireless wearable device
  for predicting septic shock in febrile patients in the emergency department:
  A machine learning-based analysis.
\newblock \emph{Sensors}, 22\penalty0 (18):\penalty0 7054, 2022.

\bibitem[Hayat et~al.(2022)Hayat, Geras, and Shamout]{hayat2022medfuse}
Nasir Hayat, Krzysztof~J Geras, and Farah~E Shamout.
\newblock Medfuse: Multi-modal fusion with clinical time-series data and chest
  x-ray images.
\newblock \emph{arXiv preprint arXiv:2207.07027}, 2022.

\bibitem[Horn et~al.(2020)Horn, Moor, Bock, Rieck, and Borgwardt]{horn2020set}
Max Horn, Michael Moor, Christian Bock, Bastian Rieck, and Karsten Borgwardt.
\newblock Set functions for time series.
\newblock In \emph{International Conference on Machine Learning}, pages
  4353--4363. PMLR, 2020.

\bibitem[Irvin et~al.(2019)Irvin, Rajpurkar, Ko, Yu, Ciurea{-}Ilcus, Chute,
  Marklund, Haghgoo, Ball, Shpanskaya, Seekins, Mong, Halabi, Sandberg, Jones,
  Larson, Langlotz, Patel, Lungren, and Ng]{DBLP:journals/corr/abs-1901-07031}
Jeremy Irvin, Pranav Rajpurkar, Michael Ko, Yifan Yu, Silviana Ciurea{-}Ilcus,
  Chris Chute, Henrik Marklund, Behzad Haghgoo, Robyn~L. Ball, Katie~S.
  Shpanskaya, Jayne Seekins, David~A. Mong, Safwan~S. Halabi, Jesse~K.
  Sandberg, Ricky Jones, David~B. Larson, Curtis~P. Langlotz, Bhavik~N. Patel,
  Matthew~P. Lungren, and Andrew~Y. Ng.
\newblock Chexpert: {A} large chest radiograph dataset with uncertainty labels
  and expert comparison.
\newblock \emph{CoRR}, abs/1901.07031, 2019.
\newblock URL \url{http://arxiv.org/abs/1901.07031}.

\bibitem[Johnson et~al.({\natexlab{a}})Johnson, Bulgarelli, Pollard, Celi,
  Mark, and Horng]{johnsonmimiced}
Alistair Johnson, Lucas Bulgarelli, Tom Pollard, Leo~Anthony Celi, Roger Mark,
  and Steven Horng.
\newblock Mimic-iv-ed.
\newblock {\natexlab{a}}.

\bibitem[Johnson et~al.({\natexlab{b}})Johnson, Pollard, Horng, Celi, and
  Mark]{johnsonmimicnote}
Alistair Johnson, Tom Pollard, Steven Horng, Leo~Anthony Celi, and Roger Mark.
\newblock Mimic-iv-note: Deidentified free-text clinical notes.
\newblock {\natexlab{b}}.

\bibitem[Johnson et~al.(2020)Johnson, Bulgarelli, Pollard, Horng, Celi, and
  Mark]{johnson2020mimic}
Alistair Johnson, Lucas Bulgarelli, Tom Pollard, Steven Horng, Leo~Anthony
  Celi, and Roger Mark.
\newblock Mimic-iv.
\newblock \emph{PhysioNet. Available online at: https://physionet.
  org/content/mimiciv/1.0/(accessed August 23, 2021)}, 2020.

\bibitem[Johnson et~al.(2019)Johnson, Pollard, Greenbaum, Lungren, Deng, Peng,
  Lu, Mark, Berkowitz, and Horng]{johnson2019mimic}
Alistair~EW Johnson, Tom~J Pollard, Nathaniel~R Greenbaum, Matthew~P Lungren,
  Chih-ying Deng, Yifan Peng, Zhiyong Lu, Roger~G Mark, Seth~J Berkowitz, and
  Steven Horng.
\newblock Mimic-cxr-jpg, a large publicly available database of labeled chest
  radiographs.
\newblock \emph{arXiv preprint arXiv:1901.07042}, 2019.

\bibitem[Kim et~al.(2019)Kim, Kim, Cho, Kim, Sol, Sung, Cho, Park, Jang, Kim,
  et~al.]{kim2019deep}
Soo~Yeon Kim, Saehoon Kim, Joongbum Cho, Young~Suh Kim, In~Suk Sol, Youngchul
  Sung, Inhyeok Cho, Minseop Park, Haerin Jang, Yoon~Hee Kim, et~al.
\newblock A deep learning model for real-time mortality prediction in
  critically ill children.
\newblock \emph{Critical care}, 23\penalty0 (1):\penalty0 1--10, 2019.

\bibitem[Kim et~al.(2021)Kim, Son, and Kim]{kim2021vilt}
Wonjae Kim, Bokyung Son, and Ildoo Kim.
\newblock Vilt: Vision-and-language transformer without convolution or region
  supervision.
\newblock In \emph{International Conference on Machine Learning}, pages
  5583--5594. PMLR, 2021.

\bibitem[Lee et~al.(2020)Lee, Yoon, Kim, Kim, Kim, So, and
  Kang]{lee2020biobert}
Jinhyuk Lee, Wonjin Yoon, Sungdong Kim, Donghyeon Kim, Sunkyu Kim, Chan~Ho So,
  and Jaewoo Kang.
\newblock Biobert: a pre-trained biomedical language representation model for
  biomedical text mining.
\newblock \emph{Bioinformatics}, 36\penalty0 (4):\penalty0 1234--1240, 2020.

\bibitem[Lee et~al.(2022)Lee, Won, Hyun, Hahn, Choi, and Lee]{lee2022self}
Kwanhyung Lee, John Won, Heejung Hyun, Sangchul Hahn, Edward Choi, and Joohyung
  Lee.
\newblock Self-supervised predictive coding and multimodal fusion advance
  patient deterioration prediction in fine-grained time resolution.
\newblock \emph{arXiv preprint arXiv:2210.16598}, 2022.

\bibitem[Liu et~al.(2021)Liu, Lin, Cao, Hu, Wei, Zhang, Lin, and
  Guo]{liu2021swin}
Ze~Liu, Yutong Lin, Yue Cao, Han Hu, Yixuan Wei, Zheng Zhang, Stephen Lin, and
  Baining Guo.
\newblock Swin transformer: Hierarchical vision transformer using shifted
  windows.
\newblock In \emph{Proceedings of the IEEE/CVF international conference on
  computer vision}, pages 10012--10022, 2021.

\bibitem[Lyu et~al.(2022)Lyu, Dong, Wong, Zheng, Abell-Hart, Wang, and
  Chen]{lyu2022multimodal}
Weimin Lyu, Xinyu Dong, Rachel Wong, Songzhu Zheng, Kayley Abell-Hart, Fusheng
  Wang, and Chao Chen.
\newblock A multimodal transformer: Fusing clinical notes with structured ehr
  data for interpretable in-hospital mortality prediction.
\newblock \emph{arXiv preprint arXiv:2208.10240}, 2022.

\bibitem[Ma et~al.(2021)Ma, Ren, Zhao, Tulyakov, Wu, and Peng]{Ma2021SMILML}
Mengmeng Ma, Jian Ren, Long Zhao, S.~Tulyakov, Cathy Wu, and Xi~Peng.
\newblock Smil: Multimodal learning with severely missing modality.
\newblock In \emph{AAAI Conference on Artificial Intelligence}, 2021.

\bibitem[Ma et~al.(2022)Ma, Ren, Zhao, Testuggine, and Peng]{ma2022multimodal}
Mengmeng Ma, Jian Ren, Long Zhao, Davide Testuggine, and Xi~Peng.
\newblock Are multimodal transformers robust to missing modality?
\newblock In \emph{Proceedings of the IEEE/CVF Conference on Computer Vision
  and Pattern Recognition}, pages 18177--18186, 2022.

\bibitem[Nagrani et~al.(2021)Nagrani, Yang, Arnab, Jansen, Schmid, and
  Sun]{Nagrani2021AttentionBF}
Arsha Nagrani, Shan Yang, Anurag Arnab, Aren Jansen, Cordelia Schmid, and Chen
  Sun.
\newblock Attention bottlenecks for multimodal fusion.
\newblock In \emph{Neural Information Processing Systems}, 2021.

\bibitem[Poklukar et~al.(2022)Poklukar, Vasco, Yin, Melo, Paiva, and
  Kragic]{Poklukar2022GeometricMC}
Petra Poklukar, Miguel Vasco, Hang Yin, Francisco~S. Melo, Ana Paiva, and
  Danica Kragic.
\newblock Geometric multimodal contrastive representation learning.
\newblock In \emph{International Conference on Machine Learning}, 2022.

\bibitem[Qiao et~al.(2019)Qiao, Wu, Ge, and Fan]{qiao2019mnn}
Zhi Qiao, Xian Wu, Shen Ge, and Wei Fan.
\newblock Mnn: multimodal attentional neural networks for diagnosis prediction.
\newblock \emph{Extraction}, 1:\penalty0 A1, 2019.

\bibitem[Qin et~al.(2021)Qin, Madan, Ratan, Karnin, Kapoor, Bhatia, and
  Kass-Hout]{qin2021improving}
Fred Qin, Vivek Madan, Ujjwal Ratan, Zohar Karnin, Vishaal Kapoor, Parminder
  Bhatia, and Taha Kass-Hout.
\newblock Improving early sepsis prediction with multi modal learning.
\newblock \emph{arXiv preprint arXiv:2107.11094}, 2021.

\bibitem[Shukla and Marlin(2019)]{shukla2019interpolation}
Satya~Narayan Shukla and Benjamin~M Marlin.
\newblock Interpolation-prediction networks for irregularly sampled time
  series.
\newblock \emph{arXiv preprint arXiv:1909.07782}, 2019.

\bibitem[Soenksen et~al.(2022)Soenksen, Ma, Zeng, Boussioux,
  Villalobos~Carballo, Na, Wiberg, Li, Fuentes, and
  Bertsimas]{soenksen2022integrated}
Luis~R Soenksen, Yu~Ma, Cynthia Zeng, Leonard Boussioux, Kimberly
  Villalobos~Carballo, Liangyuan Na, Holly~M Wiberg, Michael~L Li, Ignacio
  Fuentes, and Dimitris Bertsimas.
\newblock Integrated multimodal artificial intelligence framework for
  healthcare applications.
\newblock \emph{NPJ Digital Medicine}, 5\penalty0 (1):\penalty0 149, 2022.

\bibitem[Sung et~al.(2021)Sung, Hahn, Han, Lee, Lee, Yoo, Heo, Kim, Chung,
  et~al.]{sung2021event}
MinDong Sung, Sangchul Hahn, Chang~Hoon Han, Jung~Mo Lee, Jayoung Lee, Jinkyu
  Yoo, Jay Heo, Young~Sam Kim, Kyung~Soo Chung, et~al.
\newblock Event prediction model considering time and input error using
  electronic medical records in the intensive care unit: Retrospective study.
\newblock \emph{JMIR medical informatics}, 9\penalty0 (11):\penalty0 e26426,
  2021.

\bibitem[Suresh et~al.(2017)Suresh, Hunt, Johnson, Celi, Szolovits, and
  Ghassemi]{Suresh2017ClinicalIP}
Harini Suresh, Nathan Hunt, Alistair E.~W. Johnson, Leo~Anthony Celi, Peter
  Szolovits, and Marzyeh Ghassemi.
\newblock Clinical intervention prediction and understanding using deep
  networks.
\newblock \emph{ArXiv}, abs/1705.08498, 2017.

\bibitem[Tipirneni and Reddy(2022)]{tipirneni2022self}
Sindhu Tipirneni and Chandan~K Reddy.
\newblock Self-supervised transformer for sparse and irregularly sampled
  multivariate clinical time-series.
\newblock \emph{ACM Transactions on Knowledge Discovery from Data (TKDD)},
  16\penalty0 (6):\penalty0 1--17, 2022.

\bibitem[Tsai et~al.(2019)Tsai, Bai, Liang, Kolter, Morency, and
  Salakhutdinov]{Tsai2019MultimodalTF}
Yao-Hung~Hubert Tsai, Shaojie Bai, Paul~Pu Liang, J.~Zico Kolter,
  Louis-Philippe Morency, and Ruslan Salakhutdinov.
\newblock Multimodal transformer for unaligned multimodal language sequences.
\newblock \emph{Proceedings of the conference. Association for Computational
  Linguistics. Meeting}, 2019:\penalty0 6558--6569, 2019.

\bibitem[Vale-Silva and Rohr(2020)]{vale2020multisurv}
Lu{\'\i}s~A Vale-Silva and Karl Rohr.
\newblock Multisurv: Long-term cancer survival prediction using multimodal deep
  learning.
\newblock \emph{medRxiv}, pages 2020--08, 2020.

\bibitem[Vasco et~al.(2020)Vasco, Melo, and Paiva]{Vasco2020MHVAEAH}
Miguel Vasco, Francisco~S. Melo, and Ana Paiva.
\newblock Mhvae: a human-inspired deep hierarchical generative model for
  multimodal representation learning.
\newblock \emph{ArXiv}, abs/2006.02991, 2020.

\bibitem[Vaswani et~al.(2017)Vaswani, Shazeer, Parmar, Uszkoreit, Jones, Gomez,
  Kaiser, and Polosukhin]{Vaswani2017AttentionIA}
Ashish Vaswani, Noam~M. Shazeer, Niki Parmar, Jakob Uszkoreit, Llion Jones,
  Aidan~N. Gomez, Lukasz Kaiser, and Illia Polosukhin.
\newblock Attention is all you need.
\newblock \emph{ArXiv}, abs/1706.03762, 2017.

\bibitem[Wang and Lan(2022)]{wang2022multi}
Yifan Wang and Ying Lan.
\newblock Multi-view learning based on non-redundant fusion for icu patient
  mortality prediction.
\newblock In \emph{ICASSP 2022-2022 IEEE International Conference on Acoustics,
  Speech and Signal Processing (ICASSP)}, pages 1321--1325. IEEE, 2022.

\bibitem[Wanyan et~al.(2021)Wanyan, Honarvar, Jaladanki, Zang, Naik, Somani,
  De~Freitas, Paranjpe, Vaid, Zhang, et~al.]{wanyan2021contrastive}
Tingyi Wanyan, Hossein Honarvar, Suraj~K Jaladanki, Chengxi Zang, Nidhi Naik,
  Sulaiman Somani, Jessica~K De~Freitas, Ishan Paranjpe, Akhil Vaid, Jing
  Zhang, et~al.
\newblock Contrastive learning improves critical event prediction in covid-19
  patients.
\newblock \emph{Patterns}, 2\penalty0 (12):\penalty0 100389, 2021.

\bibitem[Zadeh et~al.(2017)Zadeh, Chen, Poria, Cambria, and
  Morency]{Zadeh2017TensorFN}
Amir Zadeh, Minghai Chen, Soujanya Poria, E.~Cambria, and Louis-Philippe
  Morency.
\newblock Tensor fusion network for multimodal sentiment analysis.
\newblock In \emph{Conference on Empirical Methods in Natural Language
  Processing}, 2017.

\bibitem[Zhang et~al.(2020)Zhang, Thadajarassiri, Sen, and
  Rundensteiner]{zhang2020time}
Dongyu Zhang, Jidapa Thadajarassiri, Cansu Sen, and Elke Rundensteiner.
\newblock Time-aware transformer-based network for clinical notes series
  prediction.
\newblock In \emph{Machine learning for healthcare conference}, pages 566--588.
  PMLR, 2020.

\end{thebibliography}
